\documentclass[letterpaper, 10 pt, conference]{ieeeconf}  

\usepackage{times}
\usepackage{epsfig}
\usepackage{graphicx}
\usepackage{amsmath}
\usepackage{amssymb}
\usepackage{soul}
\usepackage{multicol}
\usepackage{makecell}
\usepackage{multirow}
\usepackage{booktabs}
\usepackage{color, colortbl}
\usepackage{dsfont}
\usepackage{caption}
\usepackage{cite}

\definecolor{Gray}{gray}{0.9}
\newcommand{\normsq}[1]{\Big\lVert#1\Big\rVert^2_2}
\newcolumntype{L}[1]{>{\raggedright\let\newline\\\arraybackslash\hspace{0pt}}m{#1}}
\newcolumntype{C}[1]{>{\centering\let\newline\\\arraybackslash\hspace{0pt}}m{#1}}
\newcolumntype{R}[1]{>{\raggedleft\let\newline\\\arraybackslash\hspace{0pt}}m{#1}}
\usepackage[pagebackref=true,breaklinks=true,letterpaper=true,colorlinks,bookmarks=false]{hyperref}

\usepackage[capitalize]{cleveref}
\crefname{section}{Sec.}{Secs.}
\Crefname{section}{Section}{Sections}
\Crefname{table}{Table}{Tables}
\crefname{table}{Tab.}{Tabs.}

\IEEEoverridecommandlockouts                              

\definecolor{MyDarkBlue}{rgb}{0,0.08,1}
\definecolor{MyDarkGreen}{rgb}{0.02,0.6,0.02}
\definecolor{MyDarkRed}{rgb}{0.8,0.02,0.02}
\definecolor{MyDarkOrange}{rgb}{0.40,0.2,0.02}
\definecolor{MyPurple}{RGB}{111,0,255}
\definecolor{MyRed}{rgb}{1.0,0.0,0.0}
\definecolor{MyGold}{rgb}{0.75,0.6,0.12}
\definecolor{MyDarkgray}{rgb}{0.66, 0.66, 0.66}

\newcommand{\eref}[1]{Equation (\ref{#1})}

\newcommand{\tabref}[1]{Table~\ref{#1}}

\newcommand{\figref}[1]{Fig.~\ref{#1}}

\title{\LARGE \bf
HandNeRF: Learning to Reconstruct Hand-Object Interaction Scene from a Single RGB Image
}
\author{Hongsuk Choi, Nikhil Chavan-Dafle, Jiacheng Yuan, Volkan Isler, and Hyunsoo Park
\\
Samsung Research America, New York
\vspace{-4mm}
}

\begin{document}

\maketitle
\thispagestyle{empty}
\pagestyle{empty}

\begin{abstract}
This paper presents a new method to learn hand-object interaction prior for reconstructing a 3D hand-object scene from a single RGB image. 
The inference as well as training-data generation for 3D hand-object scene reconstruction is challenging due to the depth ambiguity of a single image and occlusions by the hand and object.
We turn this challenge into an opportunity by utilizing the hand shape to constrain the possible relative configuration of the hand and object geometry. We design a generalizable implicit function, HandNeRF, that explicitly encodes the correlation of the 3D hand shape features and 2D object features to predict the hand and object scene geometry. 
With experiments on real-world datasets, we show that HandNeRF can reconstruct hand-object scenes of novel grasp configurations more accurately than comparable methods. Moreover, we demonstrate that object reconstruction from HandNeRF ensures more accurate execution of downstream tasks, such as grasping and motion planning for robotic hand-over and manipulation. The code is released here: \url{https://github.com/hongsukchoi/HandNeRF_RELEASE}

%
\end{abstract}



\section{Introduction}
\label{sec:intro}

The understanding of 3D hand-object interactions, i.e., semantic reconstruction of hand and object geometry, is key to applications such as human-to-robot object handover, and augmented and virtual reality. 
Most of the current methods are primarily based on template-based approaches, where a known 3D CAD model of a hand and an object is assumed. The major focus is predicting the transformation that fits the known 3D CAD model to input observation ~\cite{tekin2019h+,hasson2020leveraging,liu2021semi}.
Even with these assumptions, the hand and object reconstruction from a single RGB image are both difficult tasks due to depth ambiguity, partial observation, and mutual occlusions.

\begin{figure}[!t]
\begin{center}
\includegraphics[width=1.0\linewidth]{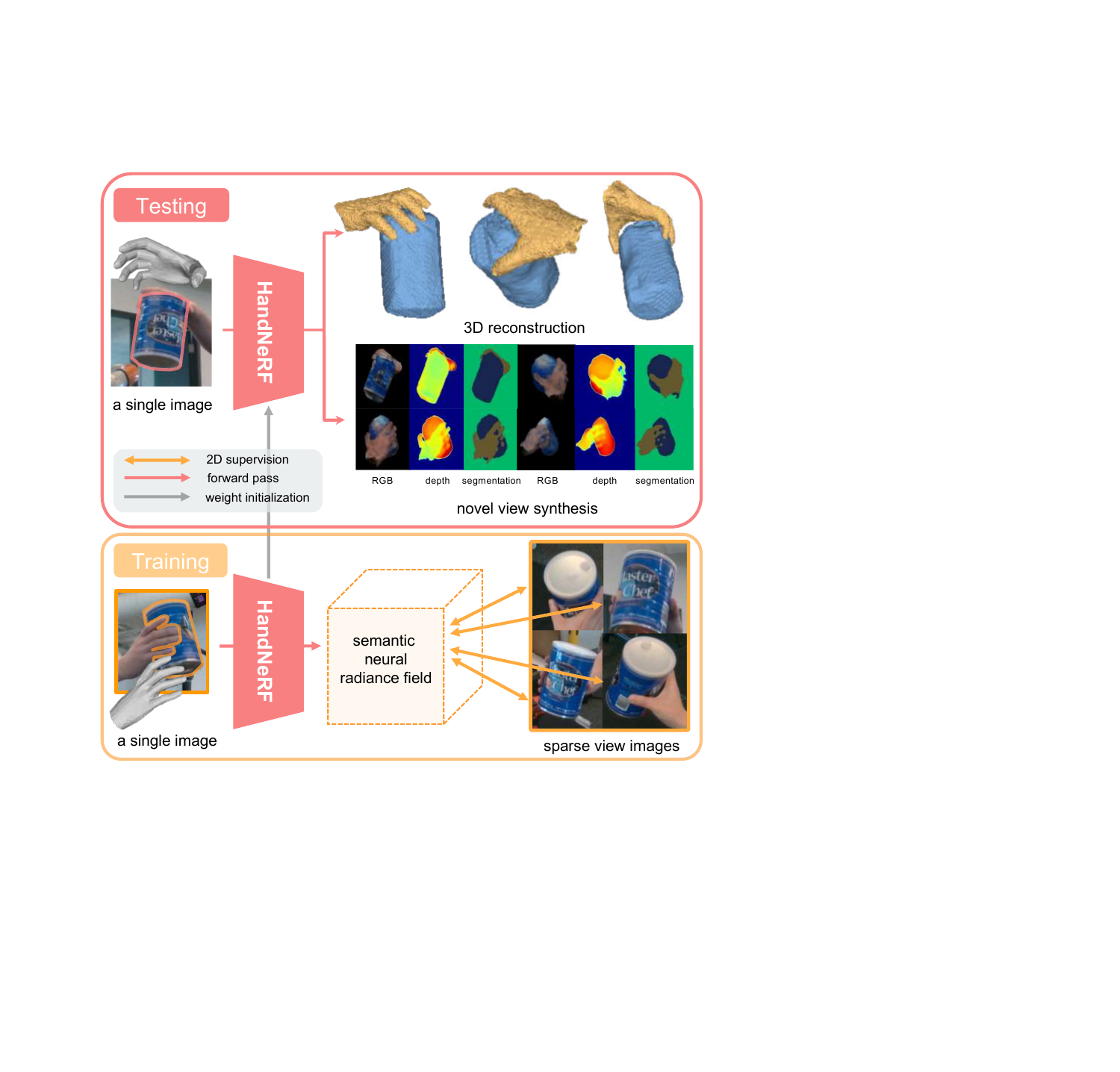}
\end{center}
\caption{\small{
Given a single RGB image of a hand-object interaction scene, HandNeRF 
predicts the hand and object's density, color, and semantics, which can be converted to reconstruction of 3D hand and object meshes and rendered to novel view images (RGB, depth, and semantic segmentation).
HandNeRF learns the correlation between hand and object geometry from different types of hand-object interactions, supervised by sparse view images.
HandNeRF is tested on a novel scene with an unseen hand-object interaction.}
}
\label{fig:teaser}
\end{figure}
%

The 3D hand reconstruction methods have seen significant advances~\cite{hampali2022keypoint,park2022handoccnet,rong2021frankmocap} due to large-scale hand datasets and automated reliable 3D  hand annotations~\cite{zimmermann2019freihand,moon2020interhand2,chao2021dexycb,zimmermann2022contrastive, pavlakos2019expressive}.
In contrast, the progress in reconstruction of grasped objects from a single RGB image is relatively limited due to lack of data.
Generating a 3D CAD model of large set of object and labeling 6D poses in hand-object interaction scenes are labor-intensive and challenging.
The sparsity of views in realworld data collection settings makes the labeling ambiguous, often requiring manual initialization and post-processing for refining 6D object pose annotations~\cite{sridhar2016real,chao2021dexycb}.


In this paper, we present a new method, named HandNeRF, that estimates a semantic neural radiance field of a hand-object interaction scene from a single RGB image and without using an object template.
HandNeRF predicts the density (occupancy), color, and semantic label (hand, object, or background) for points in the 3D space which can be used for 3D semantic reconstruction and novel view synthesis.
The key technical contribution of HandNeRF is that it alleviates the ill-posed 2D to 3D reconstruction problem by utilizing the hand shape to constrain the possible relative configuration of hand and object geometry.
In particular, HandNeRF explicitly learns the correlation between hand and object geometry to regularize their semantic reconstruction.

HandNeRF is trained on multiple hand-object interaction scenes to learn the correlation between hand and object geometry. Each scene has synchronized sparse view RGB images, 3D hand mesh annotation, and 2D semantic segmentation.
At the inference time, a single RGB image with a novel grasp configuration is given. 
\figref{fig:teaser} illustrates HandNeRF, which is trained with sparse view RGB images and generates high-quality 3D reconstruction and rendering of images from an unseen single RGB input. 
Note that we do not use depth information in the whole process, which is much more unconstrained setting for both training and testing.

We evaluate HandNeRF on realworld datasets including DexYCB~\cite{chao2021dexycb} and HO-3D v3~\cite{hampali2021ho} in terms of novel view synthesis and 3D reconstruction. 
We compare with the state-of-the-art template-free baselines~\cite{yu2021pixelnerf,ye2022s,choi2022mononhr}, which we adapt to the task of reconstructing hand-object interaction without 3D object ground truth during training. 
Following the previous works~\cite{doosti2020hope,ye2022s}, we first keep the object used in training and testing the same, but the grasp configuration at testing is chosen to be significantly different from those during training. We further evaluate the generalization capability of HandNeRF by testing the model trained on 15 DexYCB objects on 4 unseen DexYCB objects.
By learning the hand-object interaction prior with the explicit hand and object correlation, HandNeRF outperforms the baselines in generalization to novel hand grasps, which entail unseen occlusions and unseen object poses, and novel object shapes.
Furthermore, we present qualitative results demonstrating HandNeRF's ability to reconstruct objects using in-house data. 
The annotation process for this data is fully automated in a casual environment, which uses only 7 sparse view RGB cameras, without the need for 3D CAD model generation or 6D object pose labeling.
Finally, we experimentally demonstrate that HandNeRF enables more accurate execution of a downstream task of grasping for robotic hand-over and collision-free motion planning.

\section{Related Work}

Our work, HandNeRF, lies at the intersection of understanding 3D hand-object interaction and implicit neural representations. In this section, we first review the current approaches for 3D hand-object interaction reconstruction from a single RGB camera.
Then, we discuss recent methods for sparse view-specific implicit neural representations, relevant to our work.

\noindent\textbf{3D reconstruction of hand-object interaction:}
The study on the understanding of 3D hand-object interactions~\cite{rogez2015understanding,sridhar2016real,hasson2019learning,cao2021reconstructing} refers to semantic reconstruction of the hand and object geometry.
In the context of this task, the existing methods for hand and object reconstruction are primarily based on template-based approaches, where the template indicates a known 3D CAD model of a hand and an object.
The 3D hand reconstruction focuses on predicting mesh parameters, such as MANO~\cite{romero2017embodied}, and has seen a significant advance due to large-scale datasets~\cite{zimmermann2019freihand,chao2021dexycb,zimmermann2022contrastive} and success of deep learning-based approaches~\cite{ge20193d,rong2021frankmocap,moon2022accurate}.
Whereas, the 3D object reconstruction is approached as 6D pose estimation~\cite{tekin2019h+,hasson2020leveraging,liu2021semi}, which predicts the transformation that fits the known 3D CAD model to input observation.

The template-based approach for object reconstruction has two main limitations regarding collection of training data in the real world. First, it is costly and labor-intensive to obtain every object's 3D CAD model, requiring 3D laser scanning or a dense multi-view camera setup. Second, labeling 6D object poses in hand-object scenes is challenging due to hand occlusions and becomes more ambiguous if the captured views are not dense enough. In contrast, for training HandNeRF we require only a few sparse RGB views of hand-object interaction scenes and hand-pose annotations which can be automated~\cite{pavlakos2019expressive,zimmermann2022contrastive}.



Recently,~\cite{hasson2019learning,karunratanakul2020grasping,ye2022s} proposed methods that reconstruct a hand-held object without a known template.
The work of Hasson~et~al.~\cite{hasson2019learning} jointly estimated the MANO hand parameters and genus-0 object mesh by leveraging AtlasNet~\cite{groueix2018}.
Karunratanakul~et~al.~\cite{karunratanakul2020grasping} characterized the surface of the hand and object with a signed distance field.
Ye~et~al.~\cite{ye2022s} conditioned object reconstruction on the hand articulation and also predicted a signed distance field of an object.
While these methods are template-free at
the inference time, they still require 3D object meshes for
training. Therefore, they suffer with the same data collection
problems as the template-based methods.

\noindent\textbf{Implicit neural representation from sparse view RGB images:}
The sparse view-specific NeRF (Neural Radiance Field) representations have gained attention in object reconstruction~\cite{yu2021pixelnerf,wang2021ibrnet,jain2021putting,niemeyer2022regnerf} and 3D human body reconstruction~\cite{xu2021h,peng2021neural,kwon2021neural,choi2022mononhr}.
Without any 3D annotation, they reconstruct a plausible 3D scene when optimized over a single scene only with sparse views.
These methods address the limitations of multi-view reconstruction approaches such as vanilla NeRF~\cite{mildenhall2020nerf} and SfM (Structure from Motion), which require a dense capture setup and fail when given sparse views~\cite{leroy21volume}.
These representations are explored for generalization by learning object appearance and geometry priors from multiple scenes. When tested on novel scenes with unseen object poses or unseen objects, a partial 3D reconstruction is achieved, although with blurry textures and noisy geometry. This limited performance is inevitable due to sparsity of input views, but the practical applications of these methods is significant. Nevertheless, scenes with a single view or hand-held objects are are less studied. 

Our work is most relevant to the work of Choi~et~al.~\cite{choi2022mononhr}, MonoNHR.
It reconstructs a neural radiance field of a clothed human body from a single RGB image without ground truth 3D scans of clothed humans, by conditioning on a human body mesh~\cite{loper2015smpl}. 
While the task and approach are analogous to ours, MonoNHR does not explicitly encode correlation between the object (clothes) and hand (body).

\section{Method}  
\label{section:method}
\begin{figure*}[!t]
\begin{center}
\includegraphics[width=1.0\linewidth]{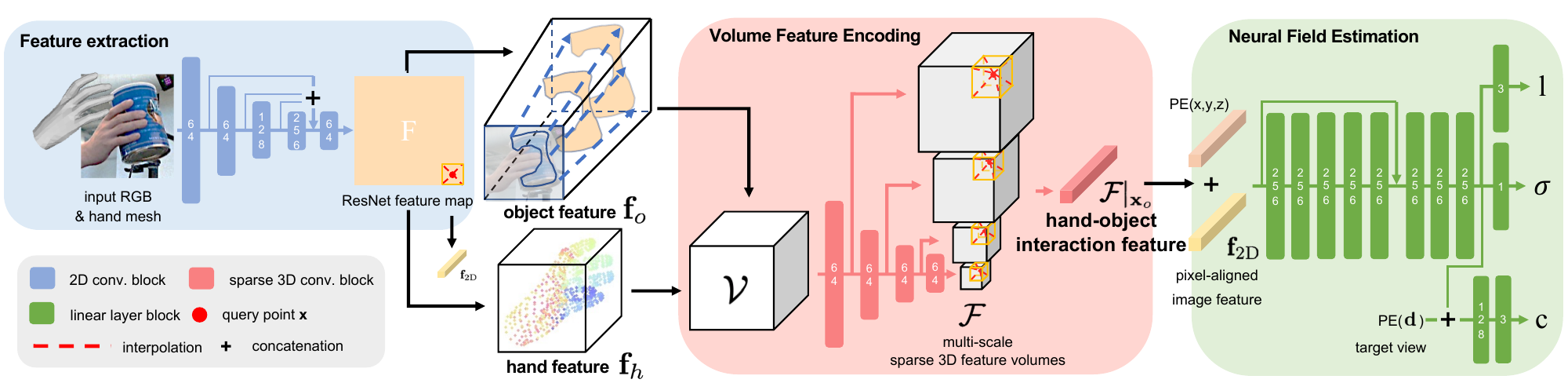}
\end{center}
\caption{
\small{
HandNeRF takes a single RGB image and predicts the volume density, color radiance, and semantic label of each query point in a neural field.} 
Different from comparable works of Ye~et~al.~\cite{ye2022s} and Choi~et~al.~\cite{choi2022mononhr} that implicitly learns the interaction between hand and object, it explicitly encodes the correlation between hand and object features in 3D space, which provides more accurate 3D reconstruction and novel view synthesis. 
}
\label{fig:HandNeRF_pipeline}
\end{figure*}

The motivation for HandNeRF is to tackle the challenges of 3D scene reconstruction from a 2D RGB image, such as depth uncertainties and partial observation.
HandNeRF achieves this by learning hand-object interaction feature that correlates the hand and object geometry, given a 3D hand shape and 2D object segmentation.
The overall pipeline of HandNeRF is depicted in \figref{fig:HandNeRF_pipeline}.
We first elaborate on the theoretical background of HandNeRF and provide the detailed implementation.

\subsection{Modeling Hand-object Interaction}

Consider a point on a 3D object, $\mathbf{x}_o\in\mathds{R}^3$ where its occupancy or density is $\sigma\in[0,1]$, i.e., one if occupied, and zero otherwise. The problem of 3D reconstruction of the object can be cast as learning a function that predicts the density given the location and associated 3D feature $\mathbf{f}_o$: 
\begin{align}
    f(\mathbf{x}_o, \mathbf{f}_o) = \sigma, \label{Eq:pifu}
\end{align}
where $f$ is an implicit function of which zero-level set defines the surface of the object. Despite the success of representing objects~\cite{yu2021pixelnerf} and humans~\cite{saito2019pifu}, \eref{Eq:pifu} has a limited capability to express complex scenes such as hand-object interaction scenes.

We extend \eref{Eq:pifu} by incorporating the interactions between the object and hand. Consider a 3D hand mesh $\mathcal{M}=\{\mathbf{m}_i\}$ that is made of a set of faces, where $\mathbf{m}_i$ is the $i^{\rm th}$ face of the mesh. Each face in the mesh is associated with a 3D feature $\mathbf{f}_h$. We marginalize the density of the object over the density predicted by the hand mesh:
\begin{align}
    f(\mathbf{x}_o, \mathbf{f}_o) = \sum_{\mathbf{x}_h\in \mathcal{M}} f(\mathbf{x}_o, \mathbf{f}_o| \mathbf{x}_h, \mathbf{f}_h) f(\mathbf{x}_h, \mathbf{f}_h), \label{Eq:handnerf_pair}
\end{align}
where $\mathbf{x}_h$ is the centroid of the vertices of the face $\mathbf{m}_i$, $f(\mathbf{x}_o, \mathbf{f}_o| \mathbf{x}_h, \mathbf{f}_h)$ is the conditional density given the hand pose and its feature, and $f(\mathbf{x}_h, \mathbf{f}_h) = \{0,1\}$ is the hand occupancy provided by 3D hand mesh estimation.

Learning $f(\mathbf{x}_o, \mathbf{f}_o| \mathbf{x}_h, \mathbf{f}_h)$ is challenging due to the quadratic complexity of pairwise relationship between all possible pairs of hand and object points $(\mathbf{x}_h, \mathbf{x}_o)$. 
Instead, we propose to learn an interaction feature $\mathcal{F}$, a correlation between $\mathbf{f}_o$ and $\mathbf{f}_h$ through a series of 3D convolutions:
\begin{align}
    \mathcal{F} = \phi_n \circ \cdots \circ \phi_1 \circ \mathcal{V}, \label{Eq:handnerf_3dcnn}
\end{align}
where $\mathcal{F} \in \mathds{R}^{w\times h \times d\times c}$ is the volume of the interaction features with $w$ width, $h$ height, $d$ depth, $c$ feature dimension, and $\phi_1,\cdots,
\phi_n$ are the 3D convolutional filters. The interaction feature  $\mathcal{F}|_{\mathbf{x}_o}$ evaluated at an object point ${\mathbf{x}_o}$ is expected to encode how hand surface points contribute to predicting the occupancy of the point ${\mathbf{x}_o}$ of the object. The input to the 3D CNN is $\mathcal{V} \in \mathds{R}^{w\times h \times d \times c^{\prime}}$, which is the feature volume with the $c^{\prime}$ feature dimension that includes both hand and object features:    \vspace{-4mm}
\begin{align}
    \mathcal{V}_{\mathbf{x}} = \left\{\begin{array}{ll}
                    \mathbf{f}_h & {\rm if~~~} \mathbf{x} \in \{\overline{\mathbf{m}}_i\}\\
                    \mathbf{f}_o & {\rm else~if~~~} \Pi \mathbf{x} \in \mathcal{O}\\
                    \mathbf{0} & {\rm otherwise}
    \end{array}\right.,\label{Eq:voxel_feature}
\end{align}
where $\mathcal{V}_\mathbf{x}$ is the feature at $\mathbf{x}$, $\{\overline{\mathbf{m}}_i\}$ is a set of the centriods of the hand mesh faces, $\Pi \mathbf{x}$ is the camera projection of $\mathbf{x}$ to the input image, and $\mathcal{O}$ is the 2D input object mask.

\begin{figure}[t]
\begin{center}
\includegraphics[width=1.0\linewidth]{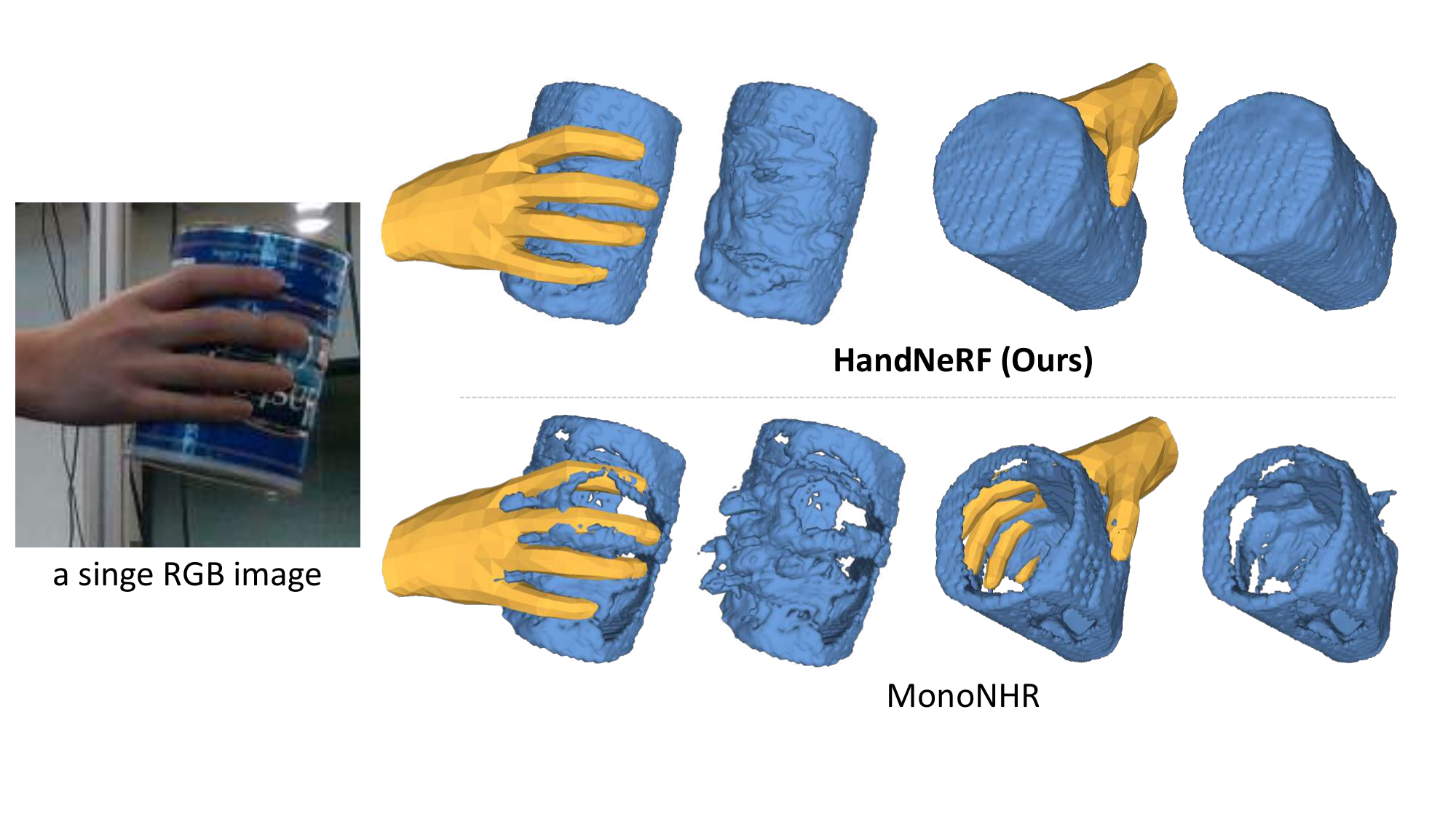}
\end{center}
\caption{
\small{We visualize object reconstruction with the hand estimation from HandOccNet~\cite{park2022handoccnet}. Using explicit hand-object interaction features, HandNeRF generates more accurate reconstruction. }
}
\label{fig:ablation1}
\end{figure}

With the interaction feature $\mathcal{F}$, we extend Equation~(\ref{Eq:pifu}) to include the color, $\mathbf{c}\in\mathds{R}^3$, and semantic label $\mathbf{l}\in[0,1]^L$ where $L=3$ is the number of semantic classes (i.e., hand, object, and background):
    \vspace{-1mm}
\begin{align}
    f(\mathbf{x}_o, \mathbf{d}, \mathbf{f}_{\rm 2D}, \mathcal{F}|_{\mathbf{x}_o}) = (\sigma, \mathbf{c}, \mathbf{l}),
\end{align}
where $\mathbf{d}$ is the rendering viewing direction, and $\mathbf{f}_{\rm 2D}$ is the pixel-aligned image feature of $\mathbf{x}_o$. With the prediction of the volume density, color radiance, and semantic label, we render each pixel with its label by integrating the density field.
Please refer to Semantic-NeRF\cite{zhi2021place} for more technical detail of semantic neural rendering.  

\figref{fig:ablation1} shows the impact of  the interaction feature $\mathcal{F}$. Our method and MonoNHR~\cite{choi2022mononhr} both use 3D CNNs to encode features volumetrically based on estimated 3D hand mesh. Unlike MonoNHR, ours explicitly learns hand-object interactions as elaborated, enabling robust object geometry reconstruction even for unobserved and occluded surfaces. 

\subsection{Implementation of HandNeRF}
We learn the representation of the hand-object interaction by minimizing the following loss: 
    \vspace{-2mm}
\begin{align}
    \mathcal{L} &= \sum\limits_{\mathbf{p} \in \mathcal{R}} \left(
    \normsq{\hat{C}(\mathbf{p}) - C(\mathbf{p})}
    - \sum_{i=1}^L L_{i}(\mathbf{p})\log(\hat{L}_{i}(\mathbf{p}))
    \right) \nonumber
\end{align}
where $\mathcal{R}$ is a set of pixels in multiview images, $\hat{C}(\mathbf{p})$ and $C(\mathbf{p})$ are the predicted and ground truth color of pixel $\mathbf{p}$, respectively, and $\hat{L}_i(\mathbf{p})$ and $L_i(\mathbf{p})$ are the predicted and ground truth semantic label at pixel $\mathbf{p}$.

We design a novel network called \textit{HandNeRF} that predicts a semantic neural radiance field from a single RGB image, as shown in \figref{fig:HandNeRF_pipeline}.
It is composed of ResNet-18~\cite{he2016deep} for feature extraction, sparse 3D convolution layers~\cite{spconv2022} for volume feature encoding, and linear layers for the neural field estimation. 
During training, the estimated semantic neural radiance field is validated by projecting on sparse views.


\noindent\textbf{Input Features:} 
We deproject a 2D image feature extracted from an image to the points in the 3D volume $\mathcal{V}$ to compose the 3D hand feature $\mathbf{f}_h$ and the 2D object feature $\mathbf{f}_o$ in \eref{Eq:voxel_feature}.
The 3D hand feature is made of three components using a MANO~\cite{romero2017embodied} hand mesh: $\mathbf{f}_h = \begin{bmatrix}\mathbf{h}^\mathsf{T}&\phi(\overline{\mathbf{m}}_i)^\mathsf{T}&\psi(i)^\mathsf{T} \end{bmatrix}^\mathsf{T}$, where $\mathbf{h}$ encodes the visual context of hand-object interaction that is obtained from the 2D image feature at the projection of $\overline{\mathbf{m}}_i$. 
$\phi(\overline{\mathbf{m}}_i)$ is a positional encoding of the centroid's coordinate, and $\psi(i)$ is the positional encoding of the face index. 
$\psi(i)$ semantically differentiates a 3D hand point $\mathbf{x}_h$ from different points that are possibly empty, belong to different hand faces, or an object.
The 2D object feature is designed in a similar fashion: $\mathbf{f}_o = \begin{bmatrix}\mathbf{o}^\mathsf{T} & \phi(\mathbf{x}_o)^\mathsf{T}& \mathbf{e}^\mathsf{T} \end{bmatrix}^\mathsf{T}$, where $\mathbf{o}$ is the 2D image feature at the projection of $\mathbf{x}_o$, and $\mathbf{e}$ is a constant value for all $\mathbf{x}_o$, where $\mathbf{x}_o \in \{\mathbf{x} \mid \Pi \mathbf{x} \in \mathcal{O} \text{ and } \mathbf{x} \notin \{\overline{\mathbf{m}}_i\}\}$.
In practice, $\psi(i)$ and $\mathbf{e}$ are randomly sampled from a Gaussian distribution and fixed during training and testing.


\begin{table*}[!t]
\small
\centering
\setlength\tabcolsep{1.0pt}
\def\arraystretch{1.1}
\caption{\small{
Ablation study. Our model M5 provides the highest rendering quality, F-scores, and lowest Chamfer Distances (CD) for novel hand-object interaction scenes' object geometry. Subscripts $_w$, $_o$, and $_h$ indicate whole, object, and hand evaluation respectively.}}
\vspace{-1mm}
\begin{tabular}{L{3.0cm}|C{1.4cm}|C{1.0cm}C{0.9cm}C{1.0cm}C{1.1cm}|C{0.9cm}C{1.1cm}C{0.9cm}|C{0.9cm}C{1.0cm}C{0.9cm}|C{0.9cm}C{1.0cm}C{0.9cm}}
\specialrule{.1em}{.05em}{.05em}
Method: features & Architect. & PSNR$\uparrow$ & IoU$\uparrow$ & SSIM$\uparrow$ & LPIPS$\downarrow$ & F$_{\text{w}}\text{-}5\uparrow$ & F$_{\text{w}}\text{-}10\uparrow$ & CD$_{\text{w}}\downarrow$ & F$_{\text{o}}\text{-}5\uparrow$ & F$_{\text{o}}\text{-}10\uparrow$ & CD$_{\text{o}}\downarrow$ & F$_{\text{h}}\text{-}5\uparrow$ & F$_{\text{h}}\text{-}10\uparrow$ & CD$_{\text{h}}\downarrow$ \\ \hline
\specialrule{.1em}{.05em}{.05em}
\cellcolor{Gray}M1: $\mathbf{f}_h,\mathbf{f}_o$+$\mathbf{f}_{\rm 2D}$ & Transf. & \cellcolor{Gray}19.40 & \cellcolor{Gray}0.61 & \cellcolor{Gray}0.63 & \cellcolor{Gray}0.30 & \cellcolor{Gray}0.36 & \cellcolor{Gray}0.64 & \cellcolor{Gray}0.85 & \cellcolor{Gray}0.32 & \cellcolor{Gray}0.56 & \cellcolor{Gray}1.57 & \cellcolor{Gray}0.27 & \cellcolor{Gray}0.55 & \cellcolor{Gray}1.42 \\ \hline
M2: $\mathbf{f}_{\rm 2D}$ & \multirow{4}{*}{3D CNN}  & 19.09 & 0.61 & 0.60 & 0.31 & 0.30 & 0.56 & 1.17 & 0.28 & 0.47 & 2.90 & 0.19 & 0.49 & 1.56 \\
\cellcolor{Gray}M3: $\mathbf{f}_h, \mathbf{f}_o$ & & \cellcolor{Gray}20.11 & \cellcolor{Gray}0.77 & \cellcolor{Gray}0.65 & \cellcolor{Gray}0.27 & \cellcolor{Gray}0.47 & \cellcolor{Gray}0.78 & \cellcolor{Gray}0.30 & \cellcolor{Gray}0.41 & \cellcolor{Gray}0.68 & \cellcolor{Gray}0.62 & \cellcolor{Gray}\textbf{0.54} & \cellcolor{Gray}\textbf{0.92} & \cellcolor{Gray}0.12 \\
M4: $\mathbf{f}_h$+$\mathbf{f}_{\rm 2D}$ & & 20.31 & 0.72 & 0.68 & 0.26 & 0.40 & 0.68 & 0.79 & 0.30 & 0.53 & 1.73 & \textbf{0.54} & \textbf{0.92} & \textbf{0.09} \\
\cellcolor{Gray}M5 (ours): $\mathbf{f}_h,\mathbf{f}_o$+$\mathbf{f}_{\rm 2D}$ & & \cellcolor{Gray}\textbf{21.66} &  \cellcolor{Gray}\textbf{0.79} & \cellcolor{Gray}\textbf{0.70} & \cellcolor{Gray}\textbf{0.24} & \cellcolor{Gray}\textbf{0.47} & \cellcolor{Gray}\textbf{0.79} & \cellcolor{Gray}\textbf{0.27} & \cellcolor{Gray}\textbf{0.43} & \cellcolor{Gray}\textbf{0.70} & \cellcolor{Gray}\textbf{0.56} & \cellcolor{Gray}0.53 & \cellcolor{Gray}0.91 & \cellcolor{Gray}0.10 \\

\specialrule{.1em}{.05em}{.05em}
\end{tabular}
\label{table:ablation_table}
\end{table*}

\noindent\textbf{3D CNN Design:}
We correlate the 3D hand feature $\mathbf{f}_h$ and the 2D object feature $\mathbf{f}_o$ with a sparse 3D CNN~\cite{spconv2022} that takes the feature volume $\mathcal{V}$ as input, to learn the interaction feature. 
$\mathcal{V}$ rasterizes 3D point coordinates in the neural radiance field with a voxel size of 5mm$\times$5mm$\times$5mm.
Before the rasterization, 3D coordinates of object points are perturbated by a random Gaussian noise during training, for augmentation. 
The sparse 3D CNN produces multi-scale feature volumes, which conceptually add up to the interaction feature volume $\mathcal{F}$ by concatenation along the feature channel dimension.
In practice, we keep the feature volumes separated and extract the interaction feature $\mathcal{F}|_{\mathbf{x}}$ of a query point $\mathbf{x}$ per volume with tri-linear interpolation.

\section{Experiments}


In this section, we first validate our design of HandNeRF method by conducting detailed ablation studies. 
Then, we evaluate against the state-of-the-art baselines~\cite{yu2021pixelnerf,ye2022s,choi2022mononhr} that are adapted to be trained on sparse view images without an object template and to be tested on a single image.
We adapted IHOI~\cite{ye2022s} to training with sparse view images, instead of template-based object annotation as in the original paper, and named it as IHOINeRF.
For IHOINeRF, training with semantic labels fails to converge where reconstruction accuracy for hand and object cannot be measured.

\noindent\textbf{Metrics:}
To assess the rendering quality, we use four metrics by comparing with the ground truth images: peak signal-to-noise ratio (PSNR), semantic segmentation intersection over union (IoU), structural similarity index (SSIM), and LPIPS~\cite{zhang2018perceptual}.
For the 3D reconstruction accuracy, we compare 3D distance with the ground truth by converting the reconstructed neural radiance field to a 3D mesh using Marching cubes algorithm~\cite{lorensen1987marching}.
F-scores at 5mm and 10mm thresholds, and Chamfer distance (CD) in millimeters are used. We evaluate hand and object separately using 3D segmentation from the predicted semantics.

\noindent\textbf{Datasets:}
We use DexYCB~\cite{chao2021dexycb} and HO-3D v3~\cite{hampali2021ho} datasets for comparison. 
In DexYCB, a hand performing object handover is captured from 8 views, and 5 sequences per object are recorded where each sequence shows a distinct grasp pattern.
Per object, we keep 4 sequences for training and 1 sequence for testing to validate generalization to novel hand grasps.
In Ho-3D v3, an object grasped in a hand is captured from 5 views and 1 sequence per object is recorded where a grasping hand pose changes over time during the sequence.
For every object, we split the data to training and testing sets such that the testing set has significantly different grasping hand poses that those in the training set.


\begin{table*}[t]
\small
\centering
\setlength\tabcolsep{1.0pt}
\def\arraystretch{1.1}
    \caption{\small{Comparison with state-of-the-art baselines on DexYCB and HO-3D v3. 
    Subscripts $_w$, $_o$, and $_h$ indicate whole, object, and hand evaluation, respectively.
    $\text{\dag}$ indicates use of ground truth 3D hand meshes for inputs,  otherwise HandOccNet~\cite{park2022handoccnet}'s estimation is used.
    MPJPEs (mean per joint position error) of the estimation are 12mm and 34mm in DexYCB and HO3D v3, respectively.}
    }
\begin{tabular}{L{1.8cm}|C{1.4cm}|C{1.0cm}C{0.9cm}C{1.0cm}C{1.1cm}|C{0.9cm}C{1.1cm}C{0.9cm}|C{0.9cm}C{1.0cm}C{0.9cm}|C{0.9cm}C{1.0cm}C{0.9cm}}
\specialrule{.1em}{.05em}{.05em}
Method & Dataset & PSNR$\uparrow$ & IoU$\uparrow$ & SSIM$\uparrow$ & LPIPS$\downarrow$ & F$_{\text{w}}\text{-}5\uparrow$ & F$_{\text{w}}\text{-}10\uparrow$ & CD$_{\text{w}}\downarrow$ & F$_{\text{o}}\text{-}5\uparrow$ & F$_{\text{o}}\text{-}10\uparrow$ & CD$_{\text{o}}\downarrow$ & F$_{\text{h}}\text{-}5\uparrow$ & F$_{\text{h}}\text{-}10\uparrow$ & CD$_{\text{h}}\downarrow$ \\ \hline
\specialrule{.1em}{.05em}{.05em}
PixelNeRF & \multirow{7}{*}{DexYCB} & 19.09 & 0.61 & 0.60 & 0.31 & 0.30 & 0.56 & 1.17 & 0.28 & 0.47 & 2.90 & 0.19 & 0.49 & 1.56 \\
\cellcolor{Gray}IHOINeRF & & \cellcolor{Gray}18.49 &  \cellcolor{Gray}- & \cellcolor{Gray}0.60 & \cellcolor{Gray}0.31 & \cellcolor{Gray}0.31 & \cellcolor{Gray}0.60 & \cellcolor{Gray}1.15 & \cellcolor{Gray}$-$ & \cellcolor{Gray}$-$ & \cellcolor{Gray}$-$ & \cellcolor{Gray}$-$ & \cellcolor{Gray}$-$ & \cellcolor{Gray}$-$ \\
IHOINeRF\dag & & 19.82 & - & 0.64 & 0.27 & 0.38 & 0.69 & 0.54 & $-$ & $-$ & $-$ & $-$ & $-$ & $-$ \\
\cellcolor{Gray}MonoNHR & & \cellcolor{Gray}19.36 & \cellcolor{Gray}0.63 & \cellcolor{Gray}0.64 & \cellcolor{Gray}0.30 & \cellcolor{Gray}0.37 & \cellcolor{Gray}0.62 & \cellcolor{Gray}3.19 &  \cellcolor{Gray}0.25 &\cellcolor{Gray}0.43 & \cellcolor{Gray}8.59 & \cellcolor{Gray}0.49 & \cellcolor{Gray}0.89 & \cellcolor{Gray}0.11 \\
MonoNHR\dag & & 19.66 & 0.68 & 0.66 & 0.29 & 0.40 & 0.64 & 3.05 & 0.26 & 0.45 & 8.58 & \textbf{0.55} & \textbf{0.92} & \textbf{0.08} \\
\cellcolor{Gray}HandNeRF & & \cellcolor{Gray}21.19 & \cellcolor{Gray}0.75 & \cellcolor{Gray}0.68 & \cellcolor{Gray}0.25 & \cellcolor{Gray}0.44 & \cellcolor{Gray}0.77 & \cellcolor{Gray}0.31 & \cellcolor{Gray}0.42 & \cellcolor{Gray}0.68 & \cellcolor{Gray}0.59 & \cellcolor{Gray}0.46 &  \cellcolor{Gray}0.88 & \cellcolor{Gray}0.12 \\
HandNeRF\dag & & \textbf{21.66} & \textbf{0.79} & \textbf{0.70} & \textbf{0.24} & \textbf{0.47} & \textbf{0.79} & \textbf{0.27} & \textbf{0.43} & \textbf{0.70} & \textbf{0.56} & 0.53 & 0.91 & 0.10\\ \hline
PixelNeRF & \multirow{7}{*}{HO3D v3} & 18.82 & 0.69 & 0.65 & 0.23 & 0.38 & 0.69 & 0.75 & 0.36 & 0.58 & 1.14 & 0.29 & 0.68 & 0.41 \\
\cellcolor{Gray}IHOINeRF & & \cellcolor{Gray}18.55 & \cellcolor{Gray}- & \cellcolor{Gray}0.65 & \cellcolor{Gray}0.23 & \cellcolor{Gray}0.28 & \cellcolor{Gray}0.56 & \cellcolor{Gray}1.15 & \cellcolor{Gray}$-$ & \cellcolor{Gray}$-$ & \cellcolor{Gray}$-$ & \cellcolor{Gray}$-$ & \cellcolor{Gray}$-$ & \cellcolor{Gray}$-$ \\
IHOINeRF\dag & & 19.40 & - & 0.68 & 0.21 & 0.41 & 0.73 & 0.65 & $-$ & $-$ & $-$ & $-$ & $-$ & $-$ \\
\cellcolor{Gray}MonoNHR & & \cellcolor{Gray}16.98 & \cellcolor{Gray}0.66 & \cellcolor{Gray}0.60 & \cellcolor{Gray}0.21 & \cellcolor{Gray}0.28 & \cellcolor{Gray}0.53 & \cellcolor{Gray}2.09 & \cellcolor{Gray}0.19 & \cellcolor{Gray}0.37 & \cellcolor{Gray}3.49 & \cellcolor{Gray}0.33 & \cellcolor{Gray}0.65 & \cellcolor{Gray}0.64 \\
MonoNHR\dag & & 19.34 & 0.75 & 0.70 & 0.22 & 0.45 & 0.74 & 0.97 & 0.36 & 0.59 & 1.50 & 0.52 & 0.92 & \textbf{0.08}  \\
\cellcolor{Gray}HandNeRF & & \cellcolor{Gray}18.04 & \cellcolor{Gray}0.68 & \cellcolor{Gray}0.63 & \cellcolor{Gray}0.23 & \cellcolor{Gray}0.38 & \cellcolor{Gray}0.70 & \cellcolor{Gray}0.35 & \cellcolor{Gray}0.40 & \cellcolor{Gray}0.66 & \cellcolor{Gray}0.41 & \cellcolor{Gray}0.45 & \cellcolor{Gray}0.51 & \cellcolor{Gray}0.78 \\
HandNeRF\dag & & \textbf{20.54} & \textbf{0.82} & \textbf{0.74} & \textbf{0.18} & \textbf{0.51} & \textbf{0.83} & \textbf{0.19} & \textbf{0.47} & \textbf{0.74} & \textbf{0.31} & \textbf{0.54} & \textbf{0.94} & \textbf{0.08} \\

\specialrule{.1em}{.05em}{.05em}
\end{tabular}
\label{table:sota_dexycb_ho3dv3}
\end{table*}

\begin{figure*}[t]
\begin{center}
\includegraphics[width=1.0\linewidth]{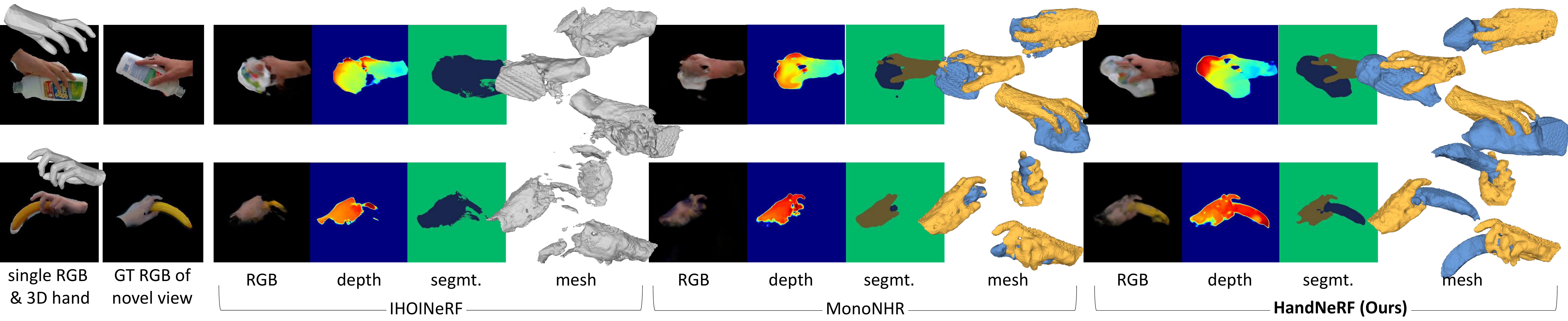}
\end{center}
\caption{
\small{
Qualitative results of novel view synthesis (image, depth, and semantic segmentation) and 3D mesh on DexYCB and HO3D v3. Ground truth hand meshes are used as input.}
}
\label{fig:sota_dexycb_and_ho3d}
\vspace{-2mm}
\end{figure*}

\begin{figure*}[t]
\begin{center}
\includegraphics[width=1.0\linewidth]{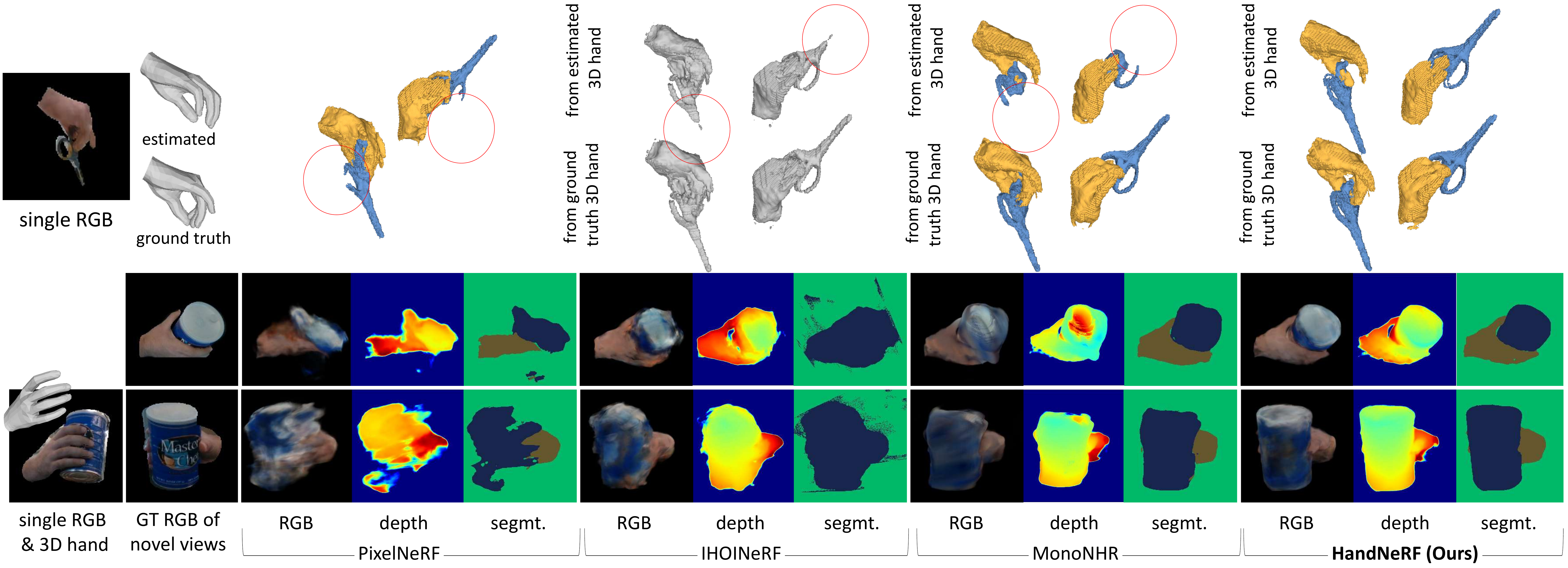}
\end{center}
\caption{
\small{
Qualitative results of novel view synthesis (image, depth, and semantic segmentation) and 3D mesh on DexYCB~\cite{chao2021dexycb} and HO3D v3~\cite{hampali2020honnotate}, given hand mesh estimation of HandOccNet~\cite{park2022handoccnet}. The bottom results for scissors are using ground truth hand mesh for reference.}
}
\vspace{-2mm}
\label{fig:sota_with_pixelnerf}
\end{figure*}

\begin{table*}[t]
\small
\centering
\setlength\tabcolsep{1.0pt}
\def\arraystretch{1.1}
    \caption{\small{Comparison with state-of-the-art baselines on unseen objects of DexYCB~\cite{chao2021dexycb}. 
    Subscripts $_w$, $_o$, and $_h$ indicate whole, object, and hand evaluation, respectively.
    The mesh estimate inputs are from HandOccNet~\cite{park2022handoccnet}. $\text{\dag}$ indicates using ground truth 3D hand meshes.}
    }
\begin{tabular}{L{1.8cm}|C{1.0cm}C{0.9cm}C{1.0cm}C{1.1cm}|C{0.9cm}C{1.1cm}C{0.9cm}|C{0.9cm}C{1.0cm}C{0.9cm}|C{0.9cm}C{1.0cm}C{0.9cm}}
\specialrule{.1em}{.05em}{.05em}
Method & PSNR$\uparrow$ & IoU$\uparrow$ & SSIM$\uparrow$ & LPIPS$\downarrow$ & F$_{\text{w}}\text{-}5\uparrow$ & F$_{\text{w}}\text{-}10\uparrow$ & CD$_{\text{w}}\downarrow$ & F$_{\text{o}}\text{-}5\uparrow$ & F$_{\text{o}}\text{-}10\uparrow$ & CD$_{\text{o}}\downarrow$ & F$_{\text{h}}\text{-}5\uparrow$ & F$_{\text{h}}\text{-}10\uparrow$ & CD$_{\text{h}}\downarrow$ \\ \hline
\specialrule{.1em}{.05em}{.05em}
PixelNeRF & 18.85 & 0.58 & 0.58 & 0.34 & 0.21 & 0.47 & 1.09 & 0.22 & 0.44 & 1.31 & 0.10 & 0.31 & 2.09 \\
\cellcolor{Gray}IHOINeRF & \cellcolor{Gray}17.89 & \cellcolor{Gray}- & \cellcolor{Gray}0.58 & \cellcolor{Gray}0.34 & \cellcolor{Gray}0.21 & \cellcolor{Gray}0.45 & \cellcolor{Gray}571.55 & \cellcolor{Gray}$-$ & \cellcolor{Gray}$-$ & \cellcolor{Gray}$-$ & \cellcolor{Gray}$-$ & \cellcolor{Gray}$-$ & \cellcolor{Gray}$-$ \\
IHOINeRF\dag & 19.94 & - & 0.64 & 0.30 & 0.43 & 0.67 & 1.03 & $-$ & $-$ & $-$ & $-$ & $-$ & $-$ \\
\cellcolor{Gray}MonoNHR & \cellcolor{Gray}17.14 & \cellcolor{Gray}0.45 & \cellcolor{Gray}0.53 & \cellcolor{Gray}0.37 & \cellcolor{Gray}0.27 & \cellcolor{Gray}0.51 & \cellcolor{Gray}1.70 &  \cellcolor{Gray}0.17 &\cellcolor{Gray}0.33 & \cellcolor{Gray}54.89 & \cellcolor{Gray}0.40 & \cellcolor{Gray}0.73 & \cellcolor{Gray}0.73 \\
MonoNHR\dag & 19.27 & 0.67 & 0.63 & 0.31 & 0.47 & 0.71 & 1.22 & 0.41 & 0.62 & 1.77 & \textbf{0.54} & \textbf{0.82} & \textbf{0.50} \\
\cellcolor{Gray}HandNeRF & \cellcolor{Gray}18.85 & \cellcolor{Gray}0.56 & \cellcolor{Gray}0.55 & \cellcolor{Gray}0.33 & \cellcolor{Gray}0.30 & \cellcolor{Gray}0.61 & \cellcolor{Gray}0.62 & \cellcolor{Gray}0.25 & \cellcolor{Gray}0.49 & \cellcolor{Gray}0.85 & \cellcolor{Gray}0.36 &  \cellcolor{Gray}0.69 & \cellcolor{Gray}0.88 \\
HandNeRF\dag & \textbf{20.83} & \textbf{0.72} & \textbf{0.66} & \textbf{0.27} & \textbf{0.51} & \textbf{0.75} & \textbf{0.72} & \textbf{0.46} & \textbf{0.68} & \textbf{1.11} & 0.51 & 0.80 & 0.52\\ \hline

\specialrule{.1em}{.05em}{.05em}
\end{tabular}
\label{table:sota_dexycb_novel_object}
\vspace*{-3mm}
\end{table*}


\begin{figure}[t]
\begin{center}
\includegraphics[width=1.0\linewidth]{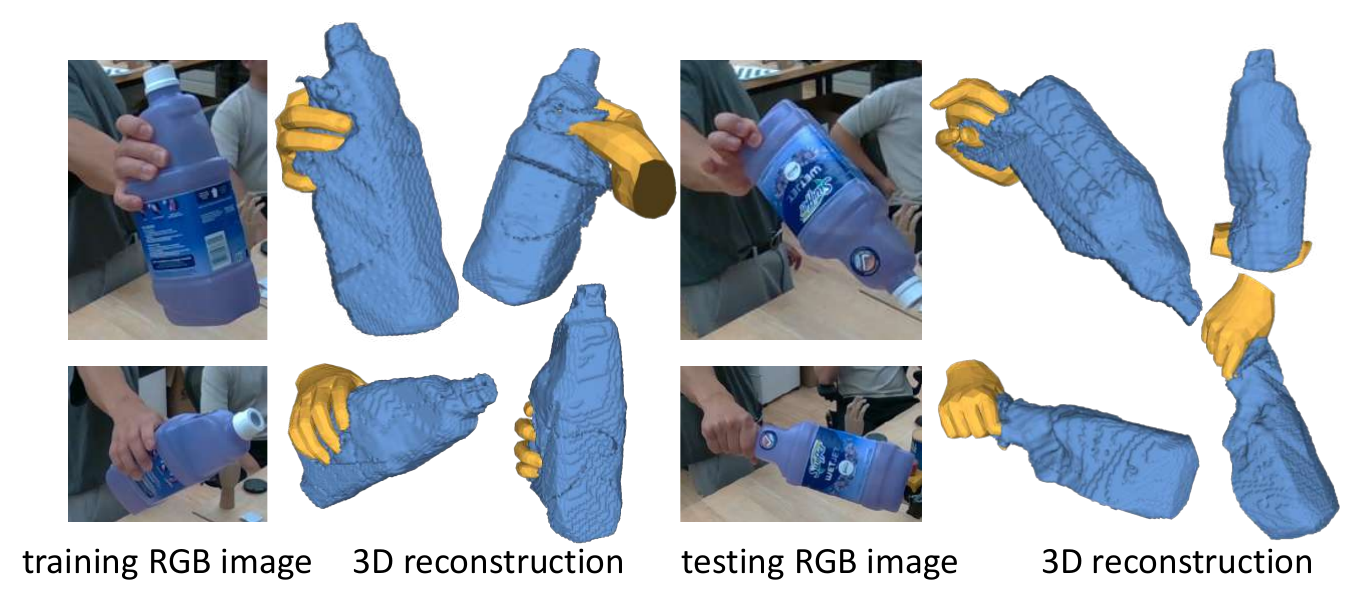}
\end{center}
\caption{\small{
HandNeRF generalizes to significantly different grasp configerations at the inference time on in-house data. The reconstructed object meshes are visualized with the input hand mesh. 
} 
}
\label{fig:inhouse_result}
\end{figure}

\begin{figure}[t]
\begin{center}
\includegraphics[width=1.0\linewidth]{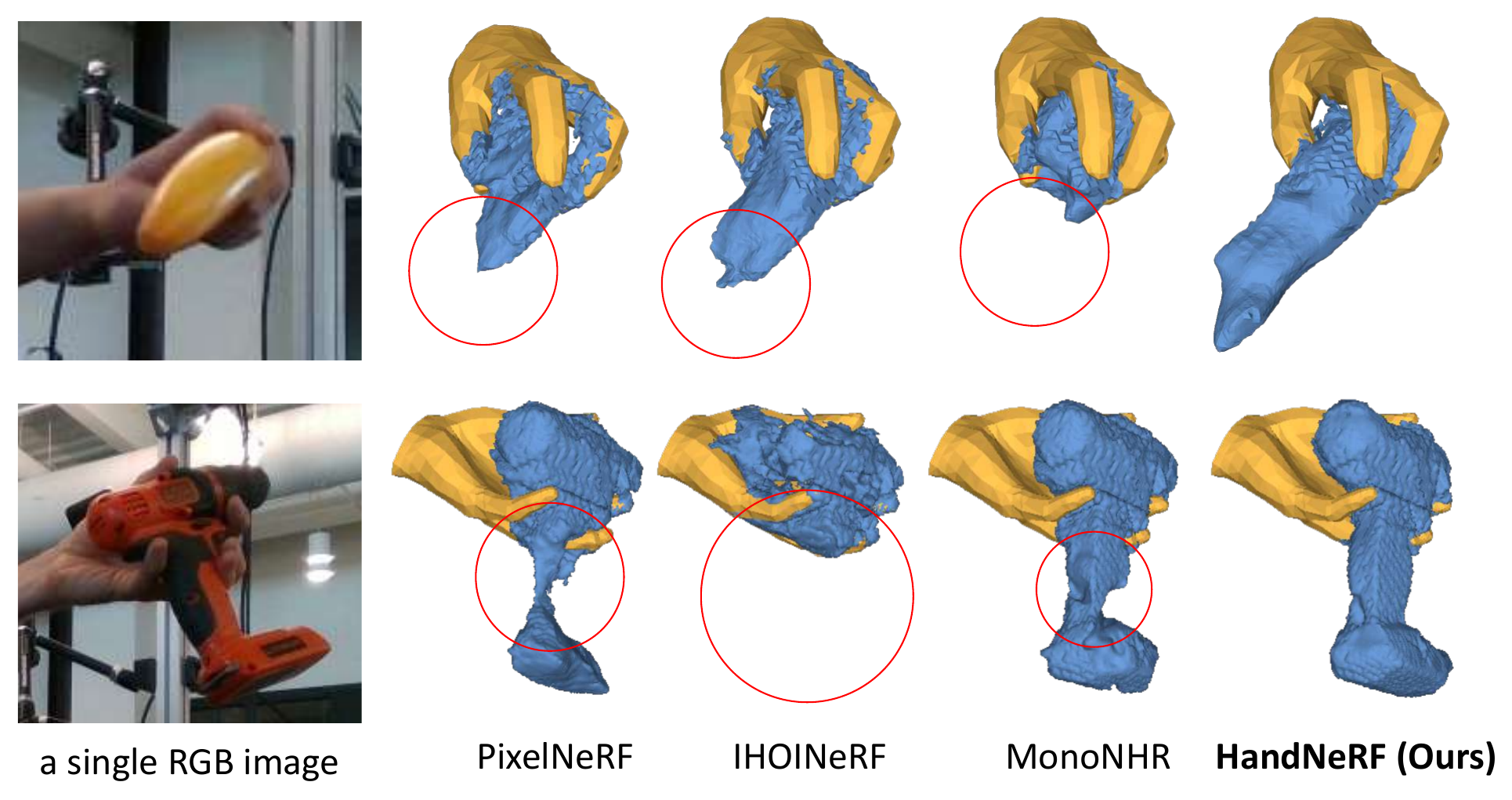}
\end{center}
\caption{
\small{
Qualitative results of generalization to novel object shapes. The reconstructed 3D object mesh is visualized along with the ground truth MANO hand mesh, which is used as input. }
}
\label{fig:sota_novel_object}
\vspace{-3mm}
\end{figure}

\begin{table}[t]
\small
\centering
\setlength\tabcolsep{1.0pt}
\def\arraystretch{1.1}
    \caption{\small{
    Downstream grasping: We compare Contact-GraspNet~\cite{sundermeyer2021contact} grasp quality on HandNeRF versus baseline reconstructions for DexYCB handover scenes. HandOccNet hand estimation is used for the methods.
}
    }
\begin{tabular}{L{1.8cm}|L{2.3cm}|C{4.2cm}}
\specialrule{.1em}{.05em}{.05em}
Input & Reconstruction & Grasp proposal success ratio$\uparrow$\\ 
\specialrule{.1em}{.05em}{.05em}
\multirow{4}{*}{RGB} & PixelNeRF & 0.46 \\
& \cellcolor{Gray}IHOINeRF & \cellcolor{Gray}0.42 \\
& MonoNHR & 0.36 \\
& \cellcolor{Gray}HandNeRF & \cellcolor{Gray}\textbf{0.63} \\ \hline
RGBD & - & 0.26 \\ \hline
GT mesh & \cellcolor{Gray}- & \cellcolor{Gray}0.77 \\
\hline

\specialrule{.1em}{.05em}{.05em}
\end{tabular}
\label{table:downstream_grasp}
\end{table}

\begin{figure}[t]
\begin{center}
\includegraphics[width=0.9\linewidth]{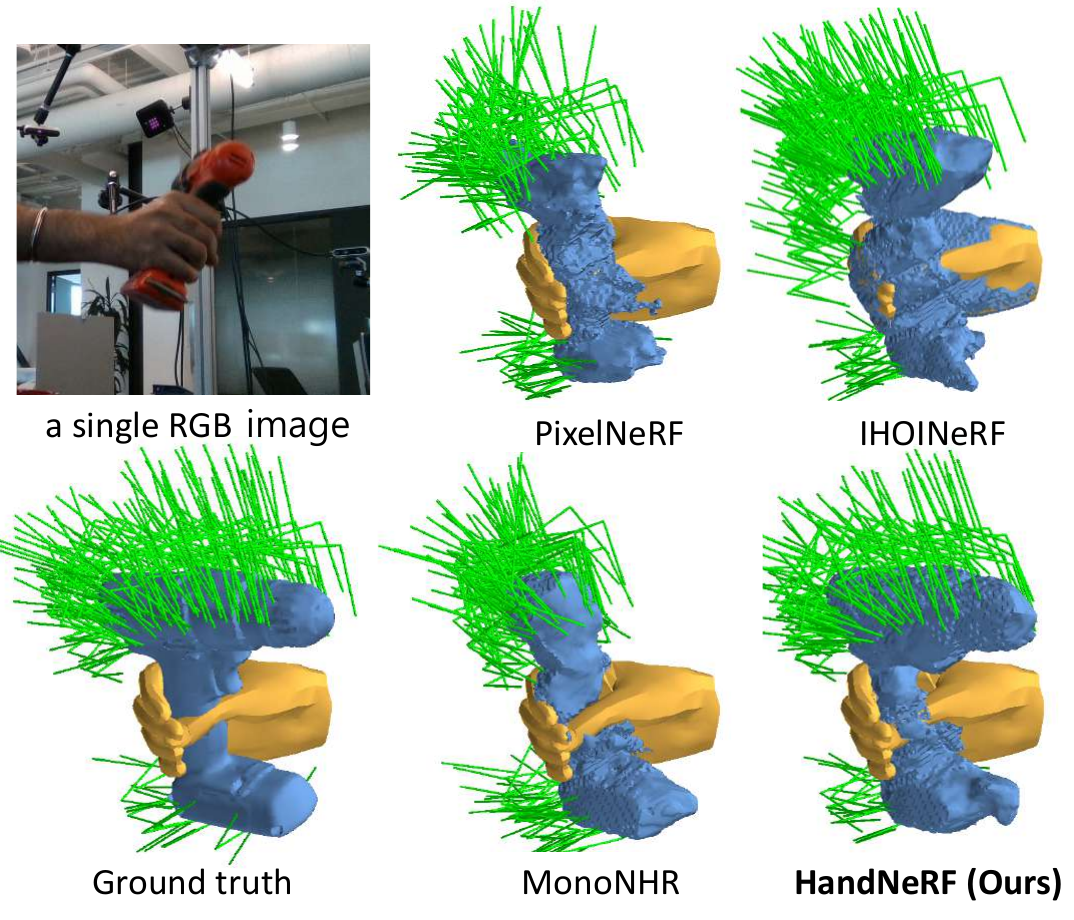}
\end{center}
\vspace{-3mm}
\caption{\small{
Qualitative results of Contact-GraspNet~\cite{sundermeyer2021contact}'s grasp proposals on reconstructed meshes of HandNeRF and the baselines. The ground truth MANO hand mesh, which is used as input and grasp collision filtering, is also visualized. 
} 
}
\label{fig:downstream_grasp}
\end{figure}

\subsection{Ablation Study}
In \tabref{table:ablation_table}, we summarize our ablation study to measure the impact of our method design choices. 
To focus on evaluation of generalization to novel grasp configurations, we train each method per object and average the metrics in the table.
We use 4 objects from DexYCB dataset with distinct shapes: `Cracker Box', `Banana', `Power Drill', and `Coffee Can'.
In inference time, all methods use the same 3D hand mesh and 2D object segmentation provided in DexYCB.

\noindent\textbf{Effect of explicit interaction encoding:}
Our main hypothesis is that learning the correlation between hand and object geometry explicitly can regulate the 3D reconstruction of the grasped object given the 3D hand shape. 
We compare two methods to validate this hypothesis: 
(M2: $\mathbf{f}_{\rm 2D}$) PixelNeRF~\cite{yu2021pixelnerf} adapted to a single input image that uses the 2D image feature without the 3D hand feature and (M3: $\mathbf{f}_h,\mathbf{f}_o$) a method that uses the 3D hand and 2D object feature without the 2D image feature. 
As shown in \tabref{table:ablation_table}, by exploiting the 3D hand feature, M3 successfully imposes constraints on the relative geometries of hand and object, and provides better generalization to novel hand-object interactions.
The results support that our learned interaction feature $\mathcal{F}$ that explicitly encodes hand-object correlations is effective to infer a 3D object geometry without requiring its 3D ground truth during training.
The performance of the 3D feature is more pronounced for our method (M5: $\mathbf{f}_h,\mathbf{f}_o+\mathbf{f}_{\rm 2D}$) that leverages both the 2D image feature and the 3D feature. 

\noindent\textbf{Significance of 2D object feature:}
HandNeRF differs from the existing approaches~\cite{ye2022s,choi2022mononhr} by using explicit representation of an object with respect to a 3D hand. 
To verify the effectiveness of the 2D object feature, we compare two methods: (M4: $\mathbf{f}_h+\mathbf{f}_{\rm 2D}$) a method that implicitly learns the hand-object interactions similar to MonoNHR~\cite{choi2022mononhr} and (M5: $\mathbf{f}_h,\mathbf{f}_o+\mathbf{f}_{\rm 2D}$) our method that explicitly models the interactions through the 2D object feature.  
As shown in \tabref{table:ablation_table}, M4 tends to overfit to hand reconstruction and produces poor results for object reconstruction.
This implies that without the explicitly defined 2D object feature and its correlation with the hand pose, a strong prior coming from the given hand pose information dominates the prediction while ignoring object information from an input image.

\noindent\textbf{Effect of 3D CNN:}
Learning hand-object interaction from Equation~\eqref{Eq:handnerf_pair} is challenging due to the complex quadratic pairwise relationship, requiring a large number of data. 
Instead, we approximate the quadratic relationship in 3D space using a 3D CNN in Equation~\eqref{Eq:handnerf_3dcnn}. 
We compare against a method (M1) that directly learns the all pairwise relationship between hand and object points using Transformer~\cite{vaswani2017attention}. 
As shown in \tabref{table:ablation_table}, using the 3D CNN (M5) outperforms M1 in all metrics. 
Considering the higher gap between training and testing PSNR of M1 (e.g., M1: 4.55  vs. M5: 3.6), the result indicates that our method based on a 3D CNN is resilient to overfitting.
Moreover, the model complexity of M1 is over 10 times than ours, comparing the number of model parameters (M1: 27.1K vs. M5: 2.1K).

\subsection{Comparison with State-of-the-art Methods}
We first evaluate the generalization to novel grasps on the objects seen during training. 
We assess the generalization to novel object shape by training on 15 DexYCB objects and testing on 4 unseen ones. Finally, we demonstrate the use of reconstruction for grasp planning for robotic handover.

\noindent\textbf{Generalization to novel grasps:}
\tabref{table:sota_dexycb_ho3dv3} and \figref{fig:sota_dexycb_and_ho3d} present quantitative and qualitative evaluation on DexYCB and HO3D v3. 
Given the ground truth grasping hand shape, HandNeRF shows the highest rendering quality scores and the highest F-scores, and the lowest CD (mm) for the whole and object geometry of the novel hand-object interaction scenes.
For example, HandNeRF achieves approximately 1.5 times higher F-scores and significantly lower CD for object reconstruction than those of PixelNeRF~\cite{yu2021pixelnerf} and MonoNHR~\cite{choi2022mononhr}.
This demonstrates HandNeRF's effective hand-object interaction priors compared to the baselines. 



We also demonstrate HandNeRF's robustness to erroneous input hand meshes in \figref{fig:sota_with_pixelnerf}, which is quantitatively verified in \tabref{table:sota_dexycb_ho3dv3}. When using the hand pose estimation from HandOccNet~\cite{park2022handoccnet}, instead of the ground truth, small pose errors in the input hand mesh significantly impact IHOINeRF and MonoNHR outputs. These methods fail to recover half the scissors, implying limitations of implicit interaction encoding. In contrast, given inaccurate 3D hand input, HandNeRF retains reasonable reconstruction quality and renders more accurate novel views far from the input. 

We further qualitatively demonstrate HandNeRF's generalization capability on in-house data in \figref{fig:inhouse_result}.
Only 7 RGB cameras are used for data collection and annotation of 3D hand mesh and 2D object segmentation is fully automated. 
Without ground truth of 3D object geometry during training, HandNeRF exhibits good generalization to significantly different grasping poses and reasonably reconstructs the grasped object from a single RGB image, leveraging the learned hand-object interaction prior.

\noindent\textbf{Generalization to novel object shape:}
As shown in \tabref{table:sota_dexycb_novel_object} and \figref{fig:sota_novel_object}, HandNeRF consistently outperforms baselines on most metrics, especially reconstructing robust object meshes despite substantial shape dissimilarity between training and test sets. Due to depth ambiguity in a single 2D RGB image, baselines fail to recover overall object shape. For instance, the `Banana' is only partially reconstructed as its length is unclear from the input. These results demonstrate HandNeRF's superior generalization, likely due to the explicit hand and object geometry encoding effectively regularizing plausible novel object geometry.


\noindent\textbf{Application to grasp planning for handover:}
We evaluate grasp proposals from Contact-GraspNet~\cite{sundermeyer2021contact} on reconstructed meshes of HandNeRF and the baselines~\cite{yu2021pixelnerf,ye2022s,choi2022mononhr}, RGBD pointcloud of the input image, and ground truth meshes. Grasps colliding with the hand mesh are filtered out before evaluation. We measure the ratio of successful grasp proposals, where a grasp is counted as successful if it envelops a part of the ground truth object mesh without colliding with it. Unseen scenes of two DexYCB objects (`Banana', `Power Drill') are used, as Contact-GraspNet performed reliably on their ground truth meshes. 


HandNeRF's object reconstruction enables a 1.5 times higher grasp proposal success ratio compared to baselines, as shown in \tabref{table:downstream_grasp}. Without depth information, HandNeRF achieves a 63\% grasp success ratio, approaching the 77\% achieved by Contact-GraspNet using ground truth meshes and far exceeding the 26\% from the input image pointcloud. \figref{fig:downstream_grasp} visually demonstrates how HandNeRF's more accurate reconstruction increases successful grasp proposals. Although the surface is locally coarse, HandNeRF's reconstructed global geometry, including the unobserved regions, enables more accurate grasp planning.




\section{Limitation and Future work}
The limitation of our method in practice is that it strongly depends on hand mesh estimation of off-the-shelf methods.
Despite the advance of the recent methods~\cite{rong2021frankmocap,park2022handoccnet,moon2022accurate}, when the hand is severely occluded by the object, the estimated mesh is not accurate enough for inferring further correlation between the hand and object geometry.
In such cases, the wrongly estimated hand information can rather heart the object reconstruction.
In the future, we will explore to integrate the hand mesh estimation into our system along with the uncertainty modeling to adjust the hand mesh's impact to the final output.

Despite outperformance, our synthesized RGB images are still blurry, when rendered from significantly different view from an input view. Inspired by recent progress on 3D scene generation via language grounding~\cite{poole2022dreamfusion}, another avenue for future research will be to leverage self-supervised perceptual supervision, such as CLIP~\cite{radford2021learning} feature consistency and object coherency. 


\section{Conclusion}

This work investigates representation learning for hand-object interactions from a single RGB image. We propose HandNeRF, a method that predicts the semantic neural radiance field of the interaction scenes. The key novelty is the utilization of hand shape to constrain the relative 3D configuration of hands and objects, encoding their correlation explicitly. Unlike existing works, HandNeRF does not require object templates for training and testing, avoiding expensive 3D labeling. Instead, it is supervised with sparse view RGB images, where conventional multi-view reconstruction methods, such as SfM (Structure from Motion), do not apply. HandNeRF outperforms state-of-the-art baselines in rendering and reconstruction on real-world data. Further, we demonstrate improved performance on downstream tasks resulting from HandNeRF's more accurate object meshes, both quantitatively and qualitatively.

\bibliographystyle{unsrt}
\bibliography{main.bbl}

\begin{thebibliography}{10}

\bibitem{tekin2019h+}
Bugra Tekin, Federica Bogo, and Marc Pollefeys.
\newblock H+ o: Unified egocentric recognition of 3d hand-object poses and
  interactions.
\newblock In {\em CVPR}, 2019.

\bibitem{hasson2020leveraging}
Yana Hasson, Bugra Tekin, Federica Bogo, Ivan Laptev, Marc Pollefeys, and
  Cordelia Schmid.
\newblock Leveraging photometric consistency over time for sparsely supervised
  hand-object reconstruction.
\newblock In {\em CVPR}, 2020.

\bibitem{liu2021semi}
Shaowei Liu, Hanwen Jiang, Jiarui Xu, Sifei Liu, and Xiaolong Wang.
\newblock Semi-supervised 3d hand-object poses estimation with interactions in
  time.
\newblock In {\em CVPR}, 2021.

\bibitem{hampali2022keypoint}
Shreyas Hampali, Sayan~Deb Sarkar, Mahdi Rad, and Vincent Lepetit.
\newblock Keypoint transformer: Solving joint identification in challenging
  hands and object interactions for accurate 3d pose estimation.
\newblock In {\em CVPR}, 2022.

\bibitem{park2022handoccnet}
JoonKyu Park, Yeonguk Oh, Gyeongsik Moon, Hongsuk Choi, and Kyoung~Mu Lee.
\newblock Handoccnet: Occlusion-robust 3d hand mesh estimation network.
\newblock In {\em CVPR}, 2022.

\bibitem{rong2021frankmocap}
Yu~Rong, Takaaki Shiratori, and Hanbyul Joo.
\newblock Frankmocap: A monocular 3d whole-body pose estimation system via
  regression and integration.
\newblock In {\em ICCV Workshops}, 2021.

\bibitem{zimmermann2019freihand}
Christian Zimmermann, Duygu Ceylan, Jimei Yang, Bryan Russell, Max Argus, and
  Thomas Brox.
\newblock Freihand: A dataset for markerless capture of hand pose and shape
  from single rgb images.
\newblock In {\em CVPR}, 2019.

\bibitem{moon2020interhand2}
Gyeongsik Moon, Shoou-I Yu, He~Wen, Takaaki Shiratori, and Kyoung~Mu Lee.
\newblock Interhand2.6m: A dataset and baseline for 3d interacting hand pose
  estimation from a single rgb image.
\newblock In {\em ECCV}, 2020.

\bibitem{chao2021dexycb}
Yu-Wei Chao, Wei Yang, Yu~Xiang, Pavlo Molchanov, Ankur Handa, Jonathan
  Tremblay, Yashraj~S Narang, Karl Van~Wyk, Umar Iqbal, Stan Birchfield, et~al.
\newblock Dexycb: A benchmark for capturing hand grasping of objects.
\newblock In {\em CVPR}, 2021.

\bibitem{zimmermann2022contrastive}
Christian Zimmermann, Max Argus, and Thomas Brox.
\newblock Contrastive representation learning for hand shape estimation.
\newblock In {\em GCPR}, 2022.

\bibitem{pavlakos2019expressive}
Georgios Pavlakos, Vasileios Choutas, Nima Ghorbani, Timo Bolkart, Ahmed~AA
  Osman, Dimitrios Tzionas, and Michael~J Black.
\newblock Expressive body capture: 3d hands, face, and body from a single
  image.
\newblock In {\em CVPR}, 2019.

\bibitem{sridhar2016real}
Srinath Sridhar, Franziska Mueller, Michael Zollh{\"o}fer, Dan Casas, Antti
  Oulasvirta, and Christian Theobalt.
\newblock Real-time joint tracking of a hand manipulating an object from rgb-d
  input.
\newblock In {\em ECCV}, 2016.

\bibitem{hampali2021ho}
Shreyas Hampali, Sayan~Deb Sarkar, and Vincent Lepetit.
\newblock Ho-3d\_v3: Improving the accuracy of hand-object annotations of the
  ho-3d dataset.
\newblock {\em arXiv}, 2021.

\bibitem{yu2021pixelnerf}
Alex Yu, Vickie Ye, Matthew Tancik, and Angjoo Kanazawa.
\newblock pixelnerf: Neural radiance fields from one or few images.
\newblock In {\em CVPR}, 2021.

\bibitem{ye2022s}
Yufei Ye, Abhinav Gupta, and Shubham Tulsiani.
\newblock What's in your hands? 3d reconstruction of generic objects in hands.
\newblock In {\em CVPR}, 2022.

\bibitem{choi2022mononhr}
Hongsuk Choi, Gyeongsik Moon, Matthieu Armando, Vincent Leroy, Kyoung~Mu Lee,
  and Gregory Rogez.
\newblock Mononhr: Monocular neural human renderer.
\newblock In {\em 3DV}, 2022.

\bibitem{doosti2020hope}
Bardia Doosti, Shujon Naha, Majid Mirbagheri, and David~J Crandall.
\newblock Hope-net: A graph-based model for hand-object pose estimation.
\newblock In {\em CVPR}, 2020.

\bibitem{rogez2015understanding}
Gr{\'e}gory Rogez, James~S Supancic, and Deva Ramanan.
\newblock Understanding everyday hands in action from rgb-d images.
\newblock In {\em ICCV}, 2015.

\bibitem{hasson2019learning}
Yana Hasson, Gul Varol, Dimitrios Tzionas, Igor Kalevatykh, Michael~J Black,
  Ivan Laptev, and Cordelia Schmid.
\newblock Learning joint reconstruction of hands and manipulated objects.
\newblock In {\em CVPR}, 2019.

\bibitem{cao2021reconstructing}
Zhe Cao, Ilija Radosavovic, Angjoo Kanazawa, and Jitendra Malik.
\newblock Reconstructing hand-object interactions in the wild.
\newblock In {\em ICCV}, 2021.

\bibitem{romero2017embodied}
Javier Romero, Dimitrios Tzionas, and Michael~J Black.
\newblock Embodied hands: Modeling and capturing hands and bodies together.
\newblock In {\em SIGGRAPH Asia}, 2017.

\bibitem{ge20193d}
Liuhao Ge, Zhou Ren, Yuncheng Li, Zehao Xue, Yingying Wang, Jianfei Cai, and
  Junsong Yuan.
\newblock 3d hand shape and pose estimation from a single rgb image.
\newblock In {\em CVPR}, 2019.

\bibitem{moon2022accurate}
Gyeongsik Moon, Hongsuk Choi, and Kyoung~Mu Lee.
\newblock Accurate 3d hand pose estimation for whole-body 3d human mesh
  estimation.
\newblock In {\em CVPR workshop}, 2022.

\bibitem{karunratanakul2020grasping}
Korrawe Karunratanakul, Jinlong Yang, Yan Zhang, Michael~J Black, Krikamol
  Muandet, and Siyu Tang.
\newblock Grasping field: Learning implicit representations for human grasps.
\newblock In {\em 3DV}, 2020.

\bibitem{groueix2018}
Thibault Groueix, Matthew Fisher, Vladimir~G. Kim, Bryan Russell, and Mathieu
  Aubry.
\newblock {AtlasNet: A Papier-M\^ach\'e Approach to Learning 3D Surface
  Generation}.
\newblock In {\em CVPR}, 2018.

\bibitem{wang2021ibrnet}
Qianqian Wang, Zhicheng Wang, Kyle Genova, Pratul~P Srinivasan, Howard Zhou,
  Jonathan~T Barron, Ricardo Martin-Brualla, Noah Snavely, and Thomas
  Funkhouser.
\newblock Ibrnet: Learning multi-view image-based rendering.
\newblock In {\em CVPR}, 2021.

\bibitem{jain2021putting}
Ajay Jain, Matthew Tancik, and Pieter Abbeel.
\newblock Putting nerf on a diet: Semantically consistent few-shot view
  synthesis.
\newblock In {\em ICCV}, 2021.

\bibitem{niemeyer2022regnerf}
Michael Niemeyer, Jonathan~T Barron, Ben Mildenhall, Mehdi~SM Sajjadi, Andreas
  Geiger, and Noha Radwan.
\newblock Regnerf: Regularizing neural radiance fields for view synthesis from
  sparse inputs.
\newblock In {\em CVPR}, 2022.

\bibitem{xu2021h}
Hongyi Xu, Thiemo Alldieck, and Cristian Sminchisescu.
\newblock H-nerf: Neural radiance fields for rendering and temporal
  reconstruction of humans in motion.
\newblock In {\em NeurIPS}, 2021.

\bibitem{peng2021neural}
Sida Peng, Yuanqing Zhang, Yinghao Xu, Qianqian Wang, Qing Shuai, Hujun Bao,
  and Xiaowei Zhou.
\newblock Neural body: Implicit neural representations with structured latent
  codes for novel view synthesis of dynamic humans.
\newblock In {\em CVPR}, 2021.

\bibitem{kwon2021neural}
Youngjoong Kwon, Dahun Kim, Duygu Ceylan, and Henry Fuchs.
\newblock {Neural Human Performer}: Learning generalizable radiance fields for
  human performance rendering.
\newblock In {\em NeurIPS}, 2021.

\bibitem{mildenhall2020nerf}
Ben Mildenhall, Pratul~P Srinivasan, Matthew Tancik, Jonathan~T Barron, Ravi
  Ramamoorthi, and Ren Ng.
\newblock Nerf: Representing scenes as neural radiance fields for view
  synthesis.
\newblock In {\em ECCV}, 2020.

\bibitem{leroy21volume}
Vincent Leroy, Jean{-}S{\'{e}}bastien Franco, and Edmond Boyer.
\newblock Volume sweeping: Learning photoconsistency for multi-view shape
  reconstruction.
\newblock {\em IJCV}, 2021.

\bibitem{loper2015smpl}
Matthew Loper, Naureen Mahmood, Javier Romero, Gerard Pons-Moll, and Michael~J
  Black.
\newblock Smpl: A skinned multi-person linear model.
\newblock {\em ACM TOG}, 2015.

\bibitem{saito2019pifu}
Shunsuke Saito, Zeng Huang, Ryota Natsume, Shigeo Morishima, Angjoo Kanazawa,
  and Hao Li.
\newblock Pifu: Pixel-aligned implicit function for high-resolution clothed
  human digitization.
\newblock In {\em ICCV}, 2019.

\bibitem{zhi2021place}
Shuaifeng Zhi, Tristan Laidlow, Stefan Leutenegger, and Andrew~J Davison.
\newblock In-place scene labelling and understanding with implicit scene
  representation.
\newblock In {\em ICCV}, 2021.

\bibitem{he2016deep}
Kaiming He, Xiangyu Zhang, Shaoqing Ren, and Jian Sun.
\newblock Deep residual learning for image recognition.
\newblock In {\em CVPR}, 2016.

\bibitem{spconv2022}
Spconv Contributors.
\newblock Spconv: Spatially sparse convolution library.
\newblock \url{https://github.com/traveller59/spconv}, 2022.

\bibitem{zhang2018perceptual}
Richard Zhang, Phillip Isola, Alexei~A Efros, Eli Shechtman, and Oliver Wang.
\newblock The unreasonable effectiveness of deep features as a perceptual
  metric.
\newblock In {\em CVPR}, 2018.

\bibitem{lorensen1987marching}
William~E Lorensen and Harvey~E Cline.
\newblock Marching cubes: A high resolution 3d surface construction algorithm.
\newblock {\em SIGGRAPH}, 1987.

\bibitem{hampali2020honnotate}
Shreyas Hampali, Mahdi Rad, Markus Oberweger, and Vincent Lepetit.
\newblock Honnotate: A method for 3d annotation of hand and object poses.
\newblock In {\em CVPR}, 2020.

\bibitem{sundermeyer2021contact}
Martin Sundermeyer, Arsalan Mousavian, Rudolph Triebel, and Dieter Fox.
\newblock Contact-graspnet: Efficient 6-dof grasp generation in cluttered
  scenes.
\newblock In {\em ICRA}, 2021.

\bibitem{vaswani2017attention}
Ashish Vaswani, Noam Shazeer, Niki Parmar, Jakob Uszkoreit, Llion Jones,
  Aidan~N Gomez, {\L}ukasz Kaiser, and Illia Polosukhin.
\newblock Attention is all you need.
\newblock In {\em NeurIPS}, 2017.

\bibitem{poole2022dreamfusion}
Ben Poole, Ajay Jain, Jonathan~T. Barron, and Ben Mildenhall.
\newblock Dreamfusion: Text-to-3d using 2d diffusion.
\newblock {\em arXiv}, 2022.

\bibitem{radford2021learning}
Alec Radford, Jong~Wook Kim, Chris Hallacy, Aditya Ramesh, Gabriel Goh,
  Sandhini Agarwal, Girish Sastry, Amanda Askell, Pamela Mishkin, Jack Clark,
  et~al.
\newblock Learning transferable visual models from natural language
  supervision.
\newblock In {\em ICML}, 2021.

\bibitem{makoviychuk2021isaac}
Viktor Makoviychuk, Lukasz Wawrzyniak, Yunrong Guo, Michelle Lu, Kier Storey,
  Miles Macklin, David Hoeller, Nikita Rudin, Arthur Allshire, Ankur Handa, and
  Gavriel State.
\newblock Isaac gym: High performance gpu-based physics simulation for robot
  learning.
\newblock {\em arXiv}, 2021.

\bibitem{tang2023rgbonly}
Zhenggang Tang, Balakumar Sundaralingam, Jonathan Tremblay, Bowen Wen, Ye~Yuan,
  Stephen Tyree, Charles Loop, Alexander Schwing, and Stan Birchfield.
\newblock Rgb-only reconstruction of tabletop scenes for collision-free
  manipulator control.
\newblock In {\em IEEE IROS}, 2023.

\bibitem{cao2017realtime}
Zhe Cao, Tomas Simon, Shih-En Wei, and Yaser Sheikh.
\newblock Realtime multi-person 2d pose estimation using part affinity fields.
\newblock In {\em CVPR}, 2017.

\bibitem{kirillov2023segany}
Alexander Kirillov, Eric Mintun, Nikhila Ravi, Hanzi Mao, Chloe Rolland, Laura
  Gustafson, Tete Xiao, Spencer Whitehead, Alexander~C. Berg, Wan-Yen Lo, Piotr
  Doll{\'a}r, and Ross Girshick.
\newblock Segment anything.
\newblock {\em arXiv}, 2023.

\bibitem{shan2020hodetector}
Dandan Shan, Jiaqi Geng, Michelle Shu, and David Fouhey.
\newblock Understanding human hands in contact at internet scale.
\newblock In {\em CVPR}, 2020.

\bibitem{kingma2015adam}
Diederik~P Kingma and Jimmy Ba.
\newblock Adam: A method for stochastic optimization.
\newblock In {\em ICLR}, 2015.

\bibitem{paszke2017automatic}
Adam Paszke, Sam Gross, Soumith Chintala, Gregory Chanan, Edward Yang, Zachary
  DeVito, Zeming Lin, Alban Desmaison, Luca Antiga, and Adam Lerer.
\newblock Automatic differentiation in pytorch.
\newblock 2017.

\end{thebibliography}

\clearpage

\twocolumn[{
\begin{center}
\begin{Large}
\textbf{\large Supplementary Material of \\ HandNeRF: Learning to Reconstruct Hand-Object Interaction Scene from a Single RGB Image}


\end{Large}
\end{center}
\vspace*{+2em}
}]

In this supplementary material, we present more experimental results that could not be included in the main manuscript due to the lack of space.

\section{Additional Qualitative Results}

\begin{figure*}[t]
\begin{center}
\includegraphics[width=0.9\linewidth]{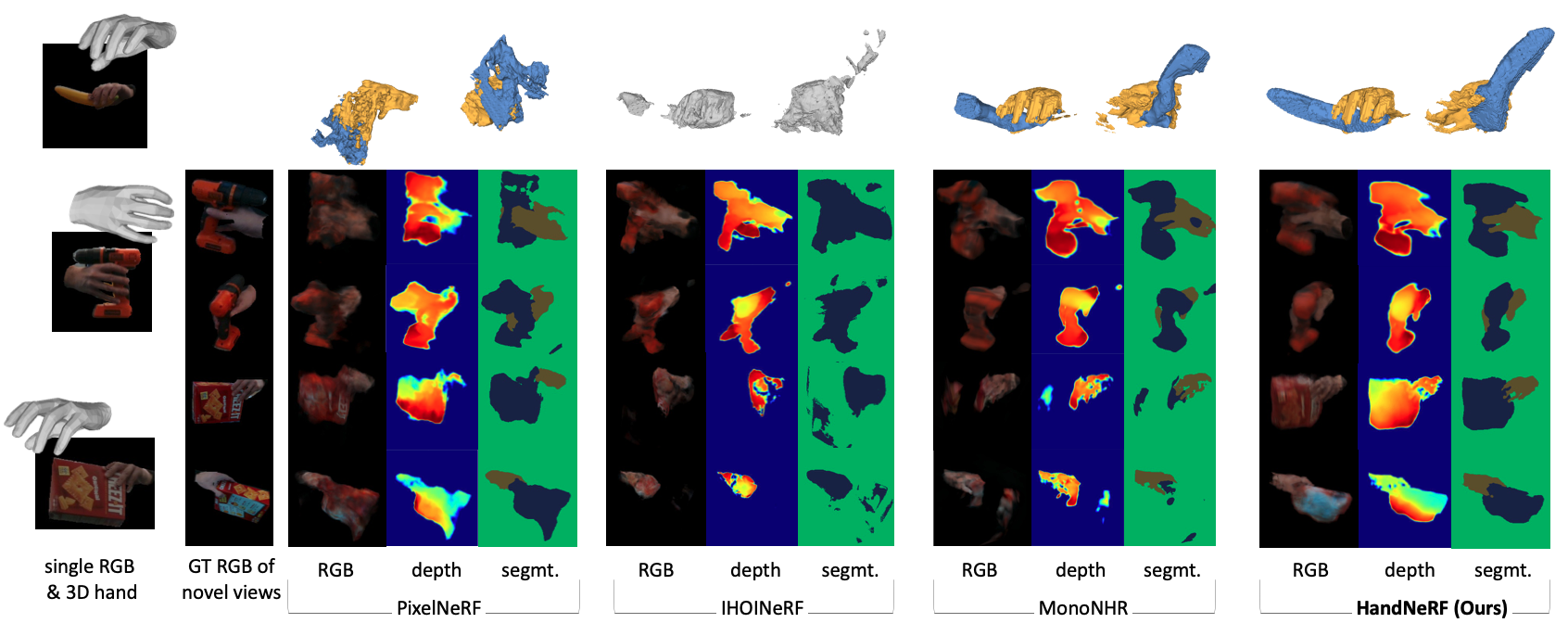}
\end{center}
\vspace{-2mm}
\caption{\small{
 Additional qualitative results. 3D hand mesh estimation of HandOccNet~\cite{park2022handoccnet} is used as input.
} 
}
\label{fig:suppl_qualitative_results}
\end{figure*}

We provide more qualitative results and comparison with baselines in \figref{fig:suppl_qualitative_results} as well as in the online video~\footnote{\url{https://youtu.be/AxkIFcymwIo}}, starting from 01:05 timestamp.
In the video, we first show more results of our HandNeRF method for novel view synthesis of RGB, depth, and semantic segmentation, and 3D reconstruction of object and hand meshes.
Then, we compare HandNeRF with PixelNeRF~\cite{yu2021pixelnerf} and IHOINeRF~\cite{ye2022s} by rendering RGB and depth images from 360 rotating views, which are significantly different from the input view.
Finally, we also demonstrate the effectiveness of HandNeRF's more accurate object reconstruction in downstream tasks such as grasp planning, collision-free motion planning, and object handover. We provide more implementation details on these tasks below.


\noindent\textbf{Collision-free Motion planning:} 
In this experiment we demonstrate that accurate object reconstruction is critical to ensure the collision-free  motion planning after the object handover to the robot. 
Specifically, in Isaac Gym simulation~\cite{makoviychuk2021isaac} environment, we compare the feasibility of the motion plans computed using object reconstructions from PixelNeRF and HandNeRF object.
We attach the reconstructed object meshes to the Panda gripper in manually selected grasp configuration. We then compute and save collision-free motions of the robot arm with the attached object reconstruction in a cluttered environment. To evaluate the feasibility of the computed arm motion, we execute the saved robot motion, but with the ground truth object mesh attached to the gripper instead of the reconstruction, similar to~\cite{tang2023rgbonly}. We monitor the motion execution for a potential collision with the environment.

\begin{figure}[t]
\begin{center}
\includegraphics[width=1.0\linewidth]{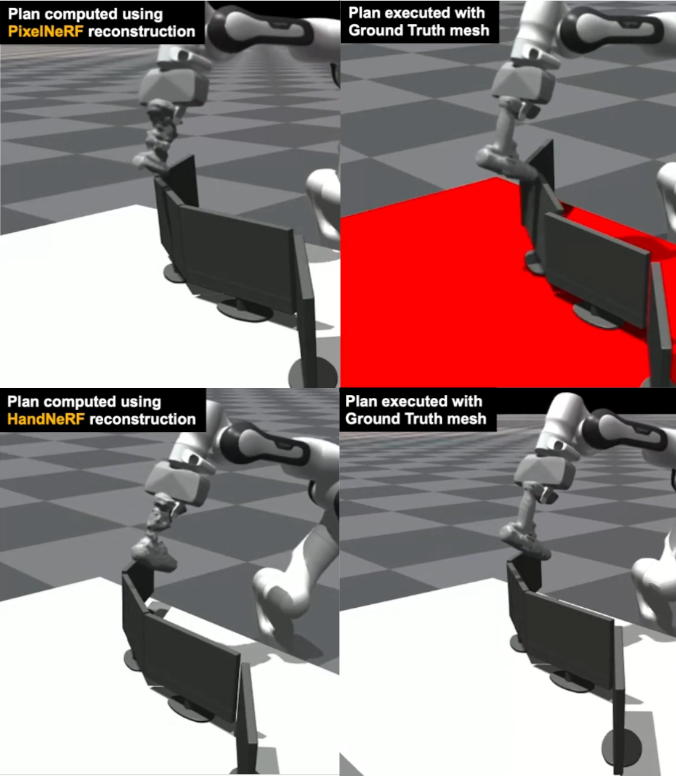}
\end{center}
\caption{\small{
 While HandNeRF's object reconstruction enables successful collision-free motion planning of the Panda Manipulation Arm, inaccurate object reconstruction of PixelNeRF leads to the execution failure. Please refer to the video.
} 
}
\label{fig:suppl_simulation}
\vspace{-2mm}
\end{figure}

As shown in \figref{fig:suppl_simulation}, we observe that the ground truth object mesh of the power drill collides with the obstacle (a computer display) in the environment when the motion plan using PixelNeRF's object reconstruction is executed. Whereas, the motion plan computed using HandNeRF does not collide with the environment. 
This motion feasibility verification with the ground truth object mesh demonstrates that the accurate object reconstruction from HandNeRF can enable collision-free motion planning in real world. 

\begin{figure}[t]
\begin{center}
\includegraphics[width=0.9\linewidth]{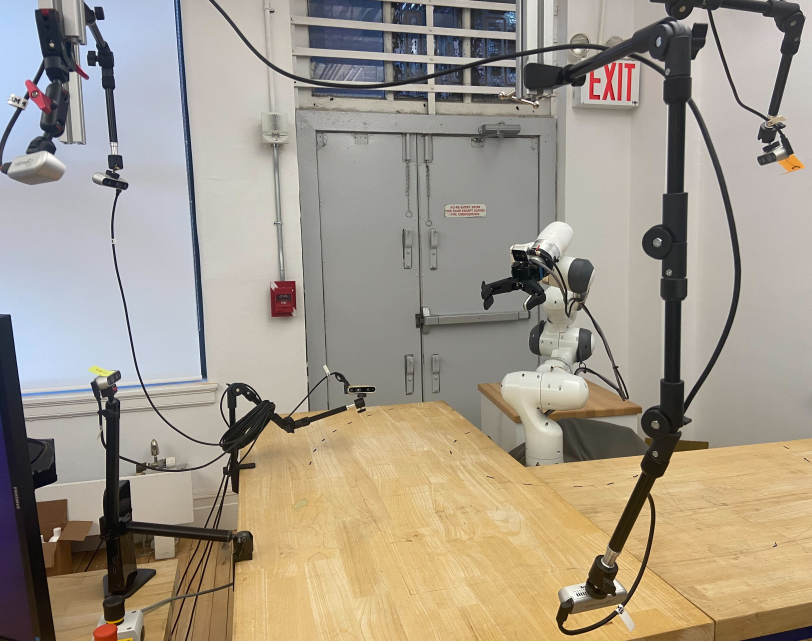}
\end{center}
\caption{\small{
 Our relatively simple scene capture environment for data collection of hand-object interaction scenes.
} 
}
\label{fig:suppl_data_collect_setting}
\end{figure}

\noindent\textbf{Realworld object handover:} We demostrate the real world handover results with `Swiffer Wetjet Refillable Bottle' object which has non-trivial shape.  
\figref{fig:suppl_data_collect_setting} shows the data collection setup. We use 7 RGB cameras for data collection; 6 RealSense D435 cameras mounted in the scene and 1 RealSense D405 camera attached to the Panda Arm, which is fixed during the capture. 
A single computer is used for synchronized capture.
Note that our data collection setting is a much more casual setting than conventional multi-view studios that require dozens of synchronized RGBD cameras that entail bandwidth issue for streaming, or involve 3D CAD models of the capturing objects.

\begin{figure}[t]
\begin{center}
\includegraphics[width=0.9\linewidth]{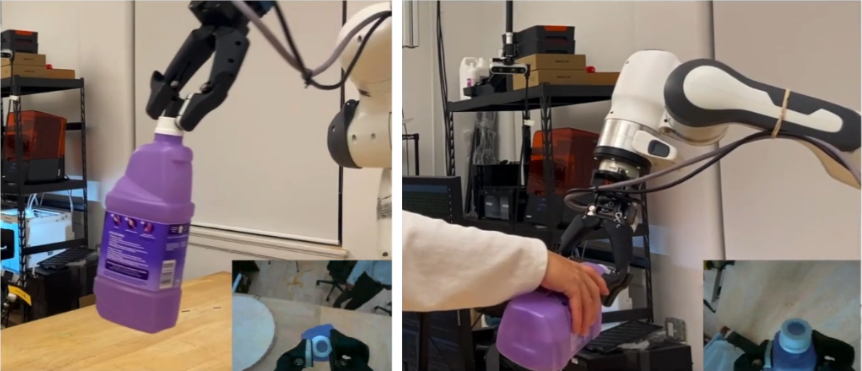}
\end{center}
\caption{\small{
We demonstrate that the reconstruction of HandNeRF from a single RGB image can used in realworld handover. Please refer to the video.
} 
}
\label{fig:suppl_real_handover}
\vspace{-2mm}
\end{figure}

Annotation of hand meshes and hand and object segmentation are fully automated. 
To obtain the 3D hand meshes, we adapted SMPLify-X~\cite{pavlakos2019expressive} to MANO hand topology and integrated multi-view 2D projection loss using OpenPose~\cite{cao2017realtime} 2D pose detection, which is implemented here \footnote{\url{https://github.com/hongsukchoi/smplify-x}}.
To obtain the segmentation, we used the publically released code of Segment-Anything~\cite{kirillov2023segany}, based on the bounding box detection of Hand Object Detector~\cite{shan2020hodetector}.

The last part of the video shows that 3D reconstruction of HandNeRF enables hand collision-free handover from a single RGB image of novel grasps and object poses in the real world, as shown in \figref{fig:suppl_real_handover}.
We extended HandNeRF to predict object segmentation for object feature encoding, and the whole process of HandNeRF took approximately 500ms on average.
The hand mesh input for HandNeRF is estimated by HandOccnet~\cite{park2022handoccnet}.
The grasp proposals are from Contact-GraspNet, and we used RRT (Rapidly-exploring random trees) motion planning.

\begin{table}[t]
\small
\centering
\setlength\tabcolsep{1.0pt}
\def\arraystretch{1.1}
    \caption{\small{
    The training and testing splits of DexYCB, regarding experiments of generalization to novel object in \tabref{table:sota_dexycb_novel_object} and \tabref{table:sota_dexycb_perobject_generalization}.}
    }
\begin{tabular}{L{2.3cm}|C{6.2cm}}
\specialrule{.1em}{.05em}{.05em}
Split & Object names\\ 
\specialrule{.1em}{.05em}{.05em}
training set & `Coffee Can', `Cracker Box', `Tomato Soup Can', `Tuna Can', `Pudding Box', `Gelatin Box', `Potted Meat Can', `Pitcher Base', `Bleach Cleanser', `Bowl', `Mug', `Wood Block', `Scissors', `Extra Large Clamp', `Foam Brick' \\\hline
testing set & `Sugar Box', `Mustard Bottle', `Banana', `Power Drill' \\
\hline
\specialrule{.1em}{.05em}{.05em}
\end{tabular}
\label{table:train_test_object}
\end{table}

\vspace*{-1em}
\begin{table}[h]
\small
\centering
\setlength\tabcolsep{1.0pt}
\def\arraystretch{1.1}
 \caption{\small{
    Comparison with the original IHOI~\cite{ye2022s}.}
    }
\begin{tabular}{C{1.5cm}|C{1.1cm}C{1.1cm}C{1.1cm}|C{1.1cm}C{1.2cm}C{1.1cm}}
\specialrule{.1em}{.1em}{.05em}
\multirow{2}{*}{Method} & \multicolumn{3}{c|}{Hand mesh estimation} &\multicolumn{3}{c}{GT hand mesh}\\
& F$_{\text{o}}\text{-}5\uparrow$ & F$_{\text{o}}\text{-}10\uparrow$ & CD$_{\text{o}}\downarrow$ & 
F$_{\text{o}}\text{-}5\uparrow$ & F$_{\text{o}}\text{-}10\uparrow$ & CD$_{\text{o}}\downarrow$ 
\\\hline
IHOI & 0.40 & 0.67 & 0.62 & 
0.65 & 0.84 & 0.25\\
Ours & 0.40 & 0.66 & 0.41 & 
0.54 & 0.74 & 0.31 \\ \hline
\specialrule{.1em}{.05em}{.05em}
\end{tabular}
\label{table:ihoi_comparison}
\vspace*{-1.4em}
\end{table}

\section{Generalization Results per Object}
We provide quantitative results per object regarding generalization to novel objects, breaking down \tabref{table:sota_dexycb_novel_object} to \tabref{table:sota_dexycb_perobject_generalization}.
The split of 15 training objects and 4 testing objects are presented in \tabref{table:train_test_object}.
We regard the generalization difficulty as `Power Drill' $=$ `Banana' $>$ `Mustard Bottle' $>$ `Sugar box', considering the difference of shape between training and testing objects.

\tabref{table:sota_dexycb_perobject_generalization} shows that HandNeRF generalizes better to more challenging object shapes, that are highly different from training object shapes, in terms of object F-scores and Chamfer distance.
For example, MonoNHR and HandNeRF are comparable in object reconstruction of `Sugar Box'. 
However, HandNeRF outperforms MonoNHR at 62.5 percentage points and 58.1 percentage points for F-score at 10mm threshold, in object reconstruction of `Banana' and `Power Drill' respectively.
The tendency is similarly observed between PixelNeRF and HandNeRF.
The results validate that HandNeRF effectively correlates the possible object geometry to the given hand shape and regularizes it, compared with the baselines.

\begin{table*}[t]
\small
\centering
\setlength\tabcolsep{1.0pt}
\def\arraystretch{1.1}
    \caption{\small{Comparison with the state-of-the-art methods for generalization to novel objects. Per-object evaluation results are presented. The outperformance of HandNeRF becomes more prominent in more challenging objects, i.e. `Banana' and `Power Drill', whose shapes are significantly different from those of training objects.}
    }
\resizebox{0.9\textwidth}{!}{%
\begin{tabular}{L{1.8cm}|C{1.4cm}|C{1.0cm}C{0.9cm}C{1.0cm}C{1.1cm}|C{0.9cm}C{1.1cm}C{1.1cm}|C{0.9cm}C{1.0cm}C{0.9cm}|C{0.9cm}C{1.0cm}C{0.9cm}}
\specialrule{.1em}{.05em}{.05em}
Method & Object & PSNR$\uparrow$ & IoU$\uparrow$ & SSIM$\uparrow$ & LPIPS$\downarrow$ & F$_{\text{w}}\text{-}5\uparrow$ & F$_{\text{w}}\text{-}10\uparrow$ & CD$_{\text{w}}\downarrow$ & F$_{\text{o}}\text{-}5\uparrow$ & F$_{\text{o}}\text{-}10\uparrow$ & CD$_{\text{o}}\downarrow$ & F$_{\text{h}}\text{-}5\uparrow$ & F$_{\text{h}}\text{-}10\uparrow$ & CD$_{\text{h}}\downarrow$ \\ \hline
\specialrule{.1em}{.05em}{.05em}
PixelNeRF & \multirow{7}{*}{\shortstack{Sugar\\Box}} & 16.83 & 0.60 & 0.55 & 0.32 & 0.22 & 0.49 & 0.83 & 0.23 & 0.48 & 0.93 & 0.12 & 0.34 & 1.86 \\
\cellcolor{Gray}IHOINeRF & & \cellcolor{Gray}15.54 &  \cellcolor{Gray}- & \cellcolor{Gray}0.52 & \cellcolor{Gray}0.36 & \cellcolor{Gray}0.26 & \cellcolor{Gray}0.52 & \cellcolor{Gray}785.03 & \cellcolor{Gray}$-$ & \cellcolor{Gray}$-$ & \cellcolor{Gray}$-$ & \cellcolor{Gray}$-$ & \cellcolor{Gray}$-$ & \cellcolor{Gray}$-$ \\
IHOINeRF\dag & & 18.52 & - & 0.63 & 0.27 & 0.42 & 0.67 & 0.55 & $-$ & $-$ & $-$ & $-$ & $-$ & $-$ \\
\cellcolor{Gray}MonoNHR & & \cellcolor{Gray}15.58 & \cellcolor{Gray}0.47 & \cellcolor{Gray}0.48 & \cellcolor{Gray}0.39 & \cellcolor{Gray}0.29 & \cellcolor{Gray}0.56 & \cellcolor{Gray}1.17 &  \cellcolor{Gray}0.20 &\cellcolor{Gray}0.40 & \cellcolor{Gray}2.23 & \cellcolor{Gray}0.44 & \cellcolor{Gray}\textbf{0.77} & \cellcolor{Gray}0.76 \\
MonoNHR\dag & & 17.95 & 0.70 & 0.58 & 0.31 & 0.47 & \textbf{0.73} & 0.51 & \textbf{0.44} & \textbf{0.67} & 0.79 & \textbf{0.50} & 0.76 & \textbf{0.69} \\
\cellcolor{Gray}HandNeRF & & \cellcolor{Gray}17.29 & \cellcolor{Gray}0.57 & \cellcolor{Gray}0.51 & \cellcolor{Gray}0.33 & \cellcolor{Gray}0.32 & \cellcolor{Gray}0.62 & \cellcolor{Gray}0.53 & \cellcolor{Gray}0.25 & \cellcolor{Gray}0.50 & \cellcolor{Gray}\textbf{0.75} & \cellcolor{Gray}0.41 &  \cellcolor{Gray}0.74 & \cellcolor{Gray}0.85 \\
HandNeRF\dag & & \textbf{19.39} & \textbf{0.75} & \textbf{0.64} & \textbf{0.26} & \textbf{0.48} & \textbf{0.73} & \textbf{0.48} & \textbf{0.44} & 0.66 & 0.96 & 0.48 & 0.74 & \textbf{0.69} \\ \hline
PixelNeRF & \multirow{7}{*}{\shortstack{Mustard\\Bottle}} & 21.15 & 0.63 & 0.56 & 0.34 & 0.22 & 0.47 & 1.13 & 0.27 & 0.52 & 0.88 & 0.07 & 0.23 & 3.10 \\
\cellcolor{Gray}IHOINeRF & & \cellcolor{Gray}19.39 & \cellcolor{Gray}- & \cellcolor{Gray}0.51 & \cellcolor{Gray}0.38 & \cellcolor{Gray}0.16 & \cellcolor{Gray}0.33 & \cellcolor{Gray}1430.62 & \cellcolor{Gray}$-$ & \cellcolor{Gray}$-$ & \cellcolor{Gray}$-$ & \cellcolor{Gray}$-$ & \cellcolor{Gray}$-$ & \cellcolor{Gray}$-$ \\
IHOINeRF\dag & & 22.00 & - & 0.60 & 0.31 & 0.37 & 0.62 & 1.49 & $-$ & $-$ & $-$ & $-$ & $-$ & $-$ \\
\cellcolor{Gray}MonoNHR & & \cellcolor{Gray}17.03 & \cellcolor{Gray}0.43 & \cellcolor{Gray}0.48 & \cellcolor{Gray}0.39 & \cellcolor{Gray}0.23 & \cellcolor{Gray}0.45 & \cellcolor{Gray}2.12 & \cellcolor{Gray}0.15 & \cellcolor{Gray}0.31 & \cellcolor{Gray}3.72 & \cellcolor{Gray}0.35 & \cellcolor{Gray}0.65 & \cellcolor{Gray}1.23 \\
MonoNHR\dag & & 20.19 & 0.74 & 0.60 & 0.29 & 0.48 & 0.73 & 1.24 & 0.45 & 0.69 & 1.76 & \textbf{0.50} & \textbf{0.75} & \textbf{1.05}  \\
\cellcolor{Gray}HandNeRF & & \cellcolor{Gray}20.27 & \cellcolor{Gray}0.57 & \cellcolor{Gray}0.51 & \cellcolor{Gray}0.36 & \cellcolor{Gray}0.27 & \cellcolor{Gray}0.56 & \cellcolor{Gray}\textbf{0.94} & \cellcolor{Gray}0.24 & \cellcolor{Gray}0.49 & \cellcolor{Gray}\textbf{1.04} & \cellcolor{Gray}0.27 & \cellcolor{Gray}0.57 & \cellcolor{Gray}1.57 \\
HandNeRF\dag & & \textbf{22.93} & \textbf{0.74} & \textbf{0.63} & \textbf{0.28} & \textbf{0.49} & \textbf{0.76} & 1.12 & 0.46 & \textbf{0.71} & 1.57 & \textbf{0.46} & 0.73 & 1.09 \\\hline
PixelNeRF & \multirow{7}{*}{Banana} & 18.88 & 0.54 & 0.66 & 0.32 & 0.20 & 0.44 & 1.44 & 0.17 & 0.32 & 2.23 & 0.12 & 0.36 & 1.54 \\
\cellcolor{Gray}IHOINeRF & & \cellcolor{Gray}19.39 & \cellcolor{Gray}- & \cellcolor{Gray}0.50 & \cellcolor{Gray}0.38 & \cellcolor{Gray}0.21 & \cellcolor{Gray}0.45 & \cellcolor{Gray}1.82 & \cellcolor{Gray}$-$ & \cellcolor{Gray}$-$ & \cellcolor{Gray}$-$ & \cellcolor{Gray}$-$ & \cellcolor{Gray}$-$ & \cellcolor{Gray}$-$ \\
IHOINeRF\dag & & 19.67 & - & 0.70 & 0.28 & 0.48 & 0.71 & 0.77 & $-$ & $-$ & $-$ & $-$ & $-$ & $-$ \\
\cellcolor{Gray}MonoNHR & & \cellcolor{Gray}17.75 & \cellcolor{Gray}0.47 & \cellcolor{Gray}0.60 & \cellcolor{Gray}0.35 & \cellcolor{Gray}0.26 & \cellcolor{Gray}0.51 & \cellcolor{Gray}1.81 & \cellcolor{Gray}0.17 & \cellcolor{Gray}0.32 & \cellcolor{Gray}259.45 & \cellcolor{Gray}0.34 & \cellcolor{Gray}0.67 & \cellcolor{Gray}0.60 \\
MonoNHR\dag & & 19.45 & 0.61 & 0.72 & 0.28 & 0.47 & 0.66 & 1.77 & 0.36 & 0.51 & 2.45 & \textbf{0.58} & \textbf{0.88} & \textbf{0.13}  \\
\cellcolor{Gray}HandNeRF & & \cellcolor{Gray}18.83 & \cellcolor{Gray}0.55 & \cellcolor{Gray}0.60 & \cellcolor{Gray}0.31 & \cellcolor{Gray}0.33 & \cellcolor{Gray}0.65 & \cellcolor{Gray}0.48 & \cellcolor{Gray}0.29 & \cellcolor{Gray}0.52 & \cellcolor{Gray}0.76 & \cellcolor{Gray}0.32 & \cellcolor{Gray}0.65 & \cellcolor{Gray}0.70 \\
HandNeRF\dag & & \textbf{20.59} & \textbf{0.70} & \textbf{0.75} & \textbf{0.25} & \textbf{0.54} & \textbf{0.76} & \textbf{0.38} & \textbf{0.49} & \textbf{0.67} & \textbf{0.52} & 0.56 & 0.87 & 0.14 \\\hline
PixelNeRF & \multirow{7}{*}{\shortstack{Power\\Drill}} & 19.16 & 0.55 & 0.58 & 0.34 & 0.20 & 0.46 & 1.07 & 0.18 & 0.40 & 1.42 & 0.10 & 0.31 & 1.77 \\
\cellcolor{Gray}IHOINeRF & & \cellcolor{Gray}18.51 & \cellcolor{Gray}- & \cellcolor{Gray}0.57 & \cellcolor{Gray}0.35 & \cellcolor{Gray}0.22 & \cellcolor{Gray}0.47 & \cellcolor{Gray}1.90 & \cellcolor{Gray}$-$ & \cellcolor{Gray}$-$ & \cellcolor{Gray}$-$ & \cellcolor{Gray}$-$ & \cellcolor{Gray}$-$ & \cellcolor{Gray}$-$ \\
IHOINeRF\dag & & 19.58 & - & 0.63 & 0.32 & 0.45 & 0.69 & 1.24 & $-$ & $-$ & $-$ & $-$ & $-$ & $-$ \\
\cellcolor{Gray}MonoNHR & & \cellcolor{Gray}18.26 & \cellcolor{Gray}0.45 & \cellcolor{Gray}0.36 & \cellcolor{Gray}0.56 & \cellcolor{Gray}0.28 & \cellcolor{Gray}0.51 & \cellcolor{Gray}1.74 & \cellcolor{Gray}0.15 & \cellcolor{Gray}0.31 & \cellcolor{Gray}3.34 & \cellcolor{Gray}0.44 & \cellcolor{Gray}0.81 & \cellcolor{Gray}0.35 \\
MonoNHR\dag & & 19.55 & 0.61 & 0.63 & 0.33 & 0.48 & 0.70 & 1.48 & 0.39 & 0.58 & 2.22 & \textbf{0.58} & \textbf{0.91} & \textbf{0.10}  \\
\cellcolor{Gray}HandNeRF & & \cellcolor{Gray}19.07 & \cellcolor{Gray}0.57 & \cellcolor{Gray}0.58 & \cellcolor{Gray}0.31 & \cellcolor{Gray}0.31 & \cellcolor{Gray}0.61 & \cellcolor{Gray}\textbf{0.52} & \cellcolor{Gray}0.24 & \cellcolor{Gray}0.49 & \cellcolor{Gray}\textbf{0.75} & \cellcolor{Gray}0.43 & \cellcolor{Gray}0.79 & \cellcolor{Gray}0.38 \\
HandNeRF\dag & & \textbf{20.52} & \textbf{0.70} & \textbf{0.65} & \textbf{0.29} & \textbf{0.52} & \textbf{0.77} & 0.82 & \textbf{0.45} & \textbf{0.67} & 1.21 & 0.54 & 0.88 & 0.14 \\ \hline

\specialrule{.1em}{.05em}{.05em}
\end{tabular}
} 
\label{table:sota_dexycb_perobject_generalization}
\end{table*}

\section{Comparison with IHOI}

We compare our HandNeRF with the original IHOI~\cite{ye2022s} that is supervised with 3D object ground truth in~\tabref{table:ihoi_comparison}.
It outperformed the previous methods~\cite{karunratanakul2020grasping,hasson2019learning} that also utilized 3D object ground truth.
We test the released model of IHOI on HO3D objects of \tabref{table:sota_dexycb_ho3dv3}.
The input hand mesh is from HandOccNet~\cite{park2022handoccnet}'s estimation and object segmentation is from the HO3D v3 annotation.

The model is expected to form an upper performance bound of our HandNeRF as it 1) exploited 3D object ground truth, 2) saw the test scenes' hand grasps and object poses during training from different views, and 3) was pre-trained on MOW dataset~\cite{cao2021reconstructing}, while our model did not.
Nonetheless, due to our explicit modeling of hand-object interactions, our method produces comparable or better accuracy with the estimated hand mesh from HandOccNet as shown in~\tabref{table:ihoi_comparison}.

\begin{table*}[t]
\small
\centering
\setlength\tabcolsep{1.0pt}
\def\arraystretch{1.1}
    \caption{\small{Ablation results of the sensitivity to the external input of HandNeRF. Without using any annotation for input, HandNeRF at the bottom row still outperforms the baselines in \tabref{table:sota_dexycb_novel_object} on DexYCB dataset.}
    }
\resizebox{0.9\linewidth}{!}{%
\begin{tabular}{L{1.8cm}|C{2.1cm}|C{1.4cm}|C{1.0cm}C{0.9cm}C{1.0cm}C{1.1cm}|C{0.9cm}C{1.1cm}C{1.1cm}|C{0.9cm}C{1.0cm}C{0.9cm}|C{0.9cm}C{1.0cm}C{0.9cm}}
\specialrule{.1em}{.05em}{.05em}
External input type & Source & Dataset & PSNR$\uparrow$ & IoU$\uparrow$ & SSIM$\uparrow$ & LPIPS$\downarrow$ & F$_{\text{w}}\text{-}5\uparrow$ & F$_{\text{w}}\text{-}10\uparrow$ & CD$_{\text{w}}\downarrow$ & F$_{\text{o}}\text{-}5\uparrow$ & F$_{\text{o}}\text{-}10\uparrow$ & CD$_{\text{o}}\downarrow$ & F$_{\text{h}}\text{-}5\uparrow$ & F$_{\text{h}}\text{-}10\uparrow$ & CD$_{\text{h}}\downarrow$ \\ \hline
\specialrule{.1em}{.05em}{.05em}
\multirow{4}{*}{Hand mesh} & Annotation & \multirow{4}{*}{HO3D v3} & 20.54 & 0.82 & 0.74 & 0.18 & 0.51 & 0.83 & 0.19 & 0.47 & 0.74 & 0.31 & 0.54 & 0.94 & 0.08 \\
& 50\% error & & 19.08 & 0.74 & 0.67 & 0.20 & 0.44 & 0.79 & 0.23 & 0.43 & 0.70 & 0.36 & 0.44 & 0.86 & 0.15 \\
& 100\% error & & 18.04 & 0.68 & 0.63 & 0.23 & 0.38 & 0.70 & 0.35 & 0.40 & 0.66 & 0.41 & 0.45 & 0.51 & 0.78 \\
& 200\% error & & 16.81 & 0.59 & 0.58 & 0.28 & 0.30 & 0.59 & 0.95 & 0.36 & 0.61 & 0.50 & 0.17 & 0.38 & 3.22 \\ \hline
\multirow{2}{*}{Segmentation} & Annotation & \multirow{2}{*}{DexYCB} & 21.66 & 0.79 & 0.70 & 0.24 & 0.47 & 0.79 & 0.27 & 0.43 & 0.70 & 0.56 & 0.53 & 0.91 & 0.10 \\
& Mask R-CNN & & 20.19 & 0.70 & 0.65 & 0.26 & 0.42 & 0.74 & 0.49 & 0.37 & 0.62 & 0.89 & 0.45 & 0.87 & 0.13 \\ \hline
\specialrule{.1em}{.05em}{.05em}
\end{tabular}
} 
\label{table:external_input_ablation}
\vspace{-2mm}
\end{table*}

\section{Sensitivity Study to External Input}
In \tabref{table:external_input_ablation}, we conduct additional experiments to evaluate our method's robustness to estimation errors of external inputs, following the work of Ye~et~al.~\cite{ye2022s}. 
For the hand mesh input, we represent the hand pose error of HandOccNet~\cite{park2022handoccnet} as an estimate deviated from the ground truth annotation.
The 100\% error source is HandOccNet's predicted hand meshes, the 50\% error source is the interpolated hand meshes of the predicted and ground truth hand meshes, and the 200\% error source is the extrapolated hand meshes of the predicted and ground truth hand meshes.  
For the object segmentation input, we evaluate our method with the MaskRCNN segmentation (object mAP: 0.873).
Without using any annotation for input, HandNeRF at the bottom row still outperforms the baselines in \tabref{table:sota_dexycb_novel_object} on DexYCB dataset.

\section{Implementation Detail}
All methods in our experimental section follow the same training and testing protocols described below.
We will release the full training and testing code along with the data.

\noindent\textbf{Training.}
We train a method 300 epochs with an initial learning rate of $10^{-3}$.
We decay the learning rate by a factor of 10 after 200th epoch.
We use the Adam~\cite{kingma2015adam} optimizer. 

For the volume rendering, we use two separate NeRF modules (i.e., fine and coarse networks) following~\cite{mildenhall2020nerf,yu2021pixelnerf}.
We render a one image from a different view per iteration and sample 1024 rays per image. 
64 points are sampled along a ray.
We gradually increase the ratio of object pixel rays to $0.5$ during training, to prevent HandNeRF from being overfitted to the hand rendering.
When an object is held by a hand, pixels of hand cover most of the input image area and tend to dominate the volume rendering without this trick.
For the feature volume of HandNeRF and MonoNHR~\cite{choi2022mononhr}, we rotate the volume by $\pi/10$ radius around XYZ axes for anti-aliasing of a 3D CNN.

We use Pytorch~\cite{paszke2017automatic} for code implementation.
We use one RTX 3090 GPU during training.
It takes approximately 45 hours to fully train HandNeRF, assuming 1000 iterations per epoch.

\noindent\textbf{Testing.}
We perform two steps to evaluate 3D reconstruction.
First, we rasterize a hand-object interacting scene with a voxel size of 2mm$\times$2mm$\times$2mm.
Then, hand and object meshes are extracted from the estimated volume densities of the neural radiance field using Marching Cubes algorithm~\cite{lorensen1987marching}, following~\cite{peng2021neural,choi2022mononhr}.
For networks that estimate a semantic label of a query point $\mathbf{x}$, we separate hand and object voxels before applying the Marching Cubes algorithm to separate hand and object meshes.
We remove small meshes with fewer than threshold voxels to reduce the effects of outliers on CD, using connected component graph theory.
The threshold is set as a 10\% of the number of input voxels.

The reconstruction time of HandNeRF, which includes the process of Marching Cubes algorithm, is 722ms (1016ms), 1015ms (1446ms), 301ms (437ms), 503ms (913ms) per scene for \textit{'002 masterchef can'}, \textit{'003 cracker box'}, \textit{'011 banana'}, \textit{'035 power drill'} of YCB objects respectively.
The numbers inside parentheses are reconstruction time with the small mesh sanitization. 
The rendering time of HandNeRF is 514ms per $3 \times 64 \times 64$ image and 8778ms per $3 \times 256 \times 256$ image.

\end{document}